\documentclass[sigconf]{acmart}
\AtBeginDocument{
  \providecommand\BibTeX{{
    \normalfont B\kern-0.5em{\scshape i\kern-0.25em b}\kern-0.8em\TeX}}}
\acmDOI{XXXXXXX.XXXXXXX}

\usepackage{url}
\usepackage{xcolor}
\usepackage[flushleft]{threeparttable}
\usepackage{algpseudocode}
\usepackage{array}
\usepackage{algorithm}
\usepackage{multirow}
\usepackage{multicol}
\usepackage{subcaption}
\usepackage{balance}
\usepackage{enumitem}
\usepackage{afterpage}

\newlength{\oldtextfloatsep}
\newcolumntype{H}{>{\setbox0=\hbox\bgroup}c<{\egroup}@{}}

\theoremstyle{definition}
\newtheorem{definition}{Definition}
\newcounter{todocounter}
\newtoggle{comments}
 \toggletrue{comments} 
\NewEnviron{doweprint}[1]{
  \iftoggle{#1}{\BODY}{}
}

\begin{document}
\title{Edge Classification on Graphs:\\
New Directions in Topological Imbalance}

\author{Xueqi Cheng}
\affiliation{
  \institution{Vanderbilt University}
  \country{}
  }
\email{xueqi.cheng@vanderbilt.edu}
\authornote{Equal contribution and co-first authors.}

\author{Yu Wang\footnotemark[1]}
\affiliation{
  \institution{Vanderbilt University}
  \country{}
  }
\email{yu.wang.1@vanderbilt.edu}

\author{Yunchao (Lance) Liu}
\affiliation{
  \institution{Vanderbilt University}
  \country{}
 }
\email{yunchao.liu@vanderbilt.edu}

\author{Yuying Zhao}
\affiliation{
  \institution{Vanderbilt University}
  \country{}
  }
\email{yuying.zhao@vanderbilt.edu}

\author{Charu C. Aggarwal}
\affiliation{
  \institution{IBM T. J. Watson Research Center}
  \country{}
  }
\email{charu@us.ibm.com}

\author{Tyler Derr}
\affiliation{
  \institution{Vanderbilt University}
  \country{}
  }
\email{tyler.derr@vanderbilt.edu}

\renewcommand{\shortauthors}{Xueqi Cheng, Yu Wang, Yunchao (Lance) Liu, Yuying Zhao, Charu C. Aggarwal, and Tyler Derr}
\renewcommand{\shorttitle}{Edge Classification on Graphs: New Directions in Topological Imbalance}

\begin{abstract}

Recent years have witnessed the remarkable success of applying Graph machine learning (GML) to node/graph classification and link prediction. However, edge classification task that enjoys numerous real-world applications such as social network analysis and cybersecurity, has not seen significant advancement with the progress of GML. To address this gap, our study pioneers a comprehensive approach to edge classification. We identify a novel `Topological Imbalance Issue', which arises from the skewed distribution of edges across different classes, affecting the local subgraph of each edge and harming the performance of edge classifications. Inspired by the recent studies in node classification that the performance discrepancy exists with varying local structural patterns, we aim to investigate if the performance discrepancy in topological imbalanced edge classification tasks can also be mitigated by characterizing the local class distribution variance. To overcome this challenge, we introduce Topological Entropy (TE), a novel topological-based metric that measures the topological imbalance for each edge. Our empirical studies confirm that TE effectively measures local class distribution variance, and indicate that prioritizing edges with high TE values can help address the issue of topological imbalance. Inspired by this observation, we develop two strategies - Topological Reweighting and TE Wedge-based Mixup - to adaptively focus training on (synthetic) edges based on their TEs. While topological reweighting directly manipulates training edge weights according to TE, our wedge-based mixup interpolates synthetic edges between high TE wedges. To further enhance performance, we integrate these strategies into a novel topological imbalance strategy for edge classification: TopoEdge. Extensive experiments on real-world datasets demonstrate the efficacy of our proposed strategies. Additionally, our curated datasets and designed experimental settings establish a new benchmark for future edge classification research, particularly in addressing imbalance issues. Our code and data are available at \href{https://github.com/XueqiC/TopoEdge}{https://github.com/XueqiC/TopoEdge}.
\end{abstract}



\received{June 2024}

\maketitle

\section{Introduction}\label{sec-introduction}

Graph Machine Learning (GML) has recently achieved unprecedented success in numerous real-world tasks over graph-structured data, including node/graph classification ~\cite{yu2020order, wu2021adapting, errica2019fair} and link prediction~\cite{zhang2018link, ai2022structure,wang2023topological}. However, the edge classification task, which is crucial for many practical applications, such as categorizing protein interactions or social media connections~\cite{kwak2010twitter, pandey2019comprehensive, jha2022prediction}, remains extensively under-explored within the GML community.

Previous approaches for edge classification in GML aim to classify edges based on the graph structure and node/edge features. Evolving from the pioneering heuristic-based edge classification method~\cite{aggarwal2016edge} which assigns edge labels based on the labels of their similar counterparts, recent work has focused on learning powerful edge embeddings to better leverage the structural and attribute information in graphs. One direction is the shallow embedding techniques~\cite{bielak2022attre2vec, wang2020edge2vec, wang2023efficient} where the random walk and deep auto-encoding methods have been adopted to capture graph structural information. Another direction lies in applying Graph Neural Network (GNN)~\cite{kipf2016semi,velivckovic2017graph, tang2019chebnet} to learn edge embeddings by either firstly learning node-level embeddings aggregated from their local subgraphs or directly generating edge embeddings~\cite{kim2019edge, gong2019exploiting} 
Despite their effectiveness, 
they all neglect the imbalance in edge classification, which could significantly compromise the classification performance and restrict their practical utility. For example, failure to recognize fraudulent behaviors in e-commerce networks, which are much less commonly seen, may result in enormous monetary losses.

\begin{figure}
     \centering
     \begin{subfigure}[b]{0.225\textwidth}
         \centering
         \includegraphics[width=.91\textwidth]{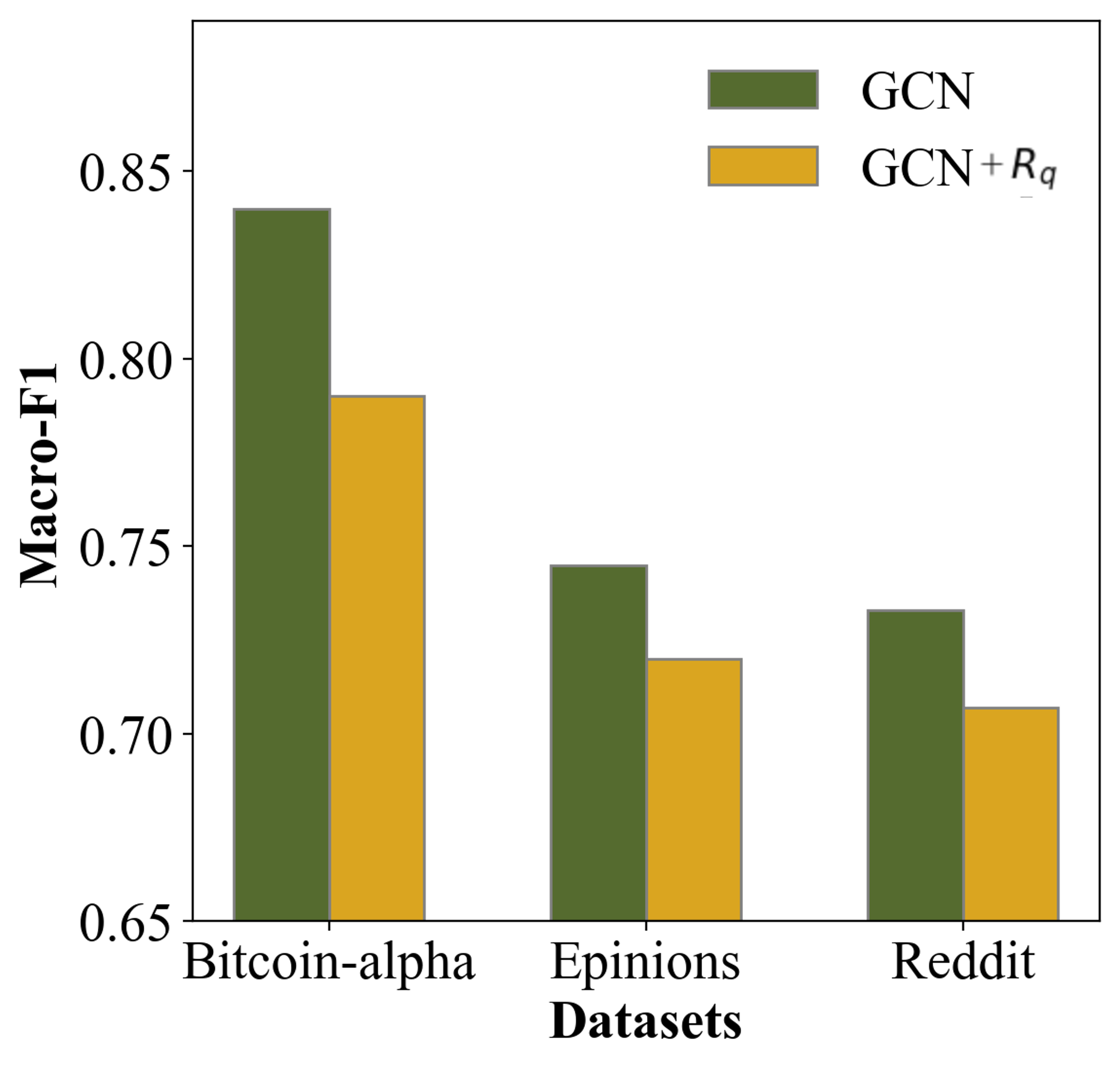}
         \vskip -1ex
         \caption{}
         \label{fig-gcn_qr}
     \end{subfigure}
     \begin{subfigure}[b]{0.225\textwidth}
         \centering
         \includegraphics[width=.91\textwidth]{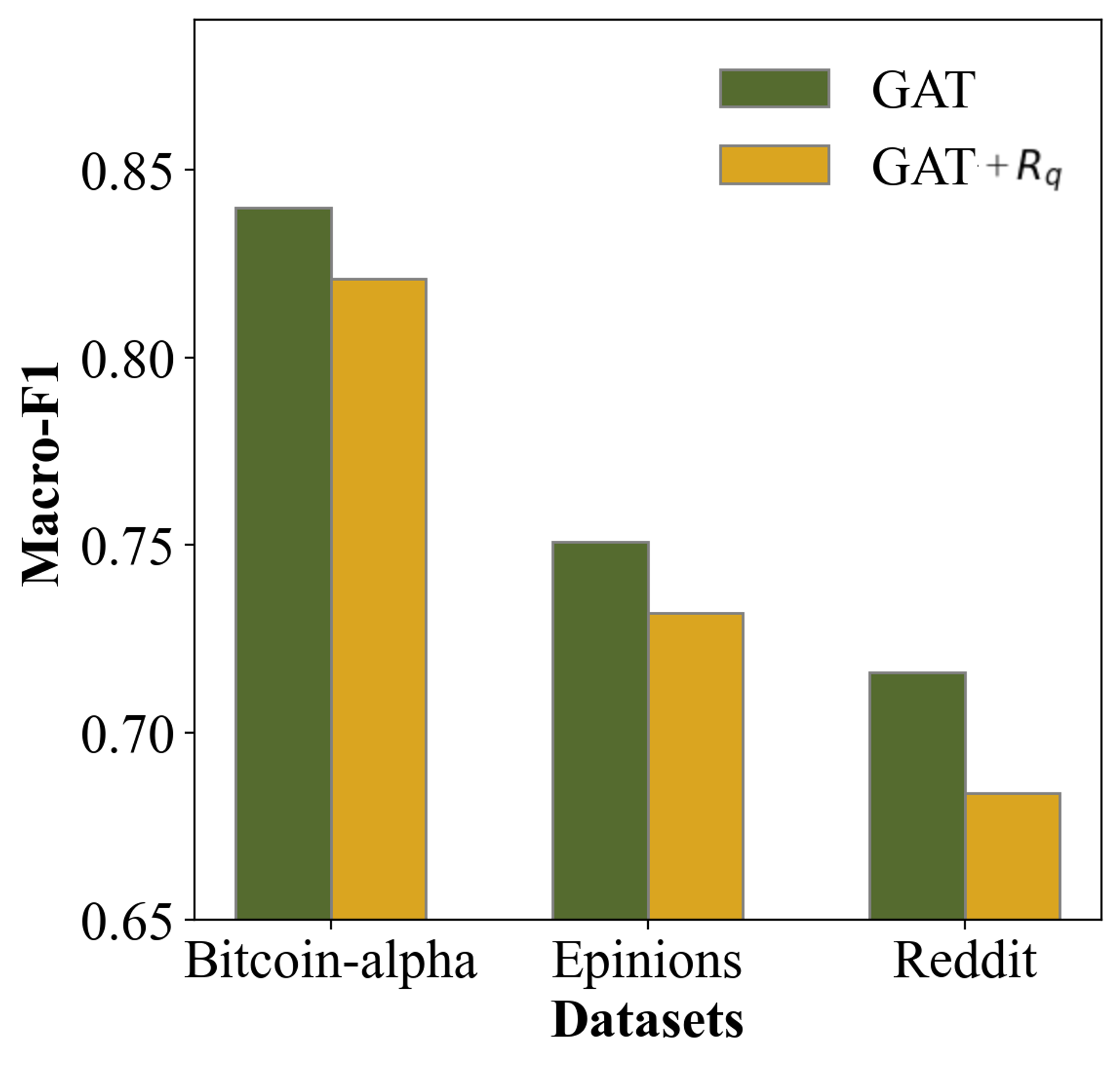}
         \vskip -1ex
         \caption{}
         \label{fig-gat_qr}
     \end{subfigure}
     \vskip -2ex
     \caption{After equipping GNNs with off-the-shelf quantity reweight ($R_q$), 
     GCN achieves worse overall Macro-F1 shown in \textbf{(a)} and similarly observed with GAT in \textbf{(b)}. 
     }
     \label{fig-intro}
     \vskip -2ex
\end{figure}

Even though numerous strategies have been developed for addressing the imbalance problem, most of them are designed for general machine learning tasks~\cite{guo2008class, chawla2002smote} or specifically for node/graph classification~\cite{wang2022imbalanced, zhao2021graphsmote,wang2021distance} in GML with no tailored modification towards edge classification. As seen in Figure~\ref{fig-intro}, applying one of the most well-established imbalance mitigation strategies, quantity reweighting, surprisingly results in inferior performance on three datasets across two GNN backbones to the original one. This motivates us to go beyond considering the edge imbalance issue merely from the quantitative perspective and more focus on the topological perspective. Actually, in node classification, previous works~\cite{zhu2021graph, mao2023demystifying} have discovered a strong correlation between the local topology pattern of a node and its GNNs' classification performance, which leads to the performance discrepancy among nodes with varying local topology. In parallel, research in imbalanced node/graph classification~\cite{chen2021topology, song2022tam, wang2022imbalanced} also discovered a novel imbalance at the topology level and found its close correlation to the node/graph classification performance. Arising from the surprising findings in Figure~\ref{fig-intro} and further based on experiences from prior works~\cite{zhu2021graph, mao2023demystifying, chen2021topology, song2022tam, wang2022imbalanced}, we hypothesize that the topology imbalance issue also exists in edge classification tasks and edges with varying local topology patterns would also exhibit varying classification performance.

To verify the above hypothesis, we first propose a metric, Topological Entropy (TE), to quantify the topological imbalance for edges based on the label distribution of incident edges in their local subgraphs. After empirically verifying the correlation between the training difficulty of edges and their TE values (Figure~\ref{edge_cat1}-\ref{topo-tracc}), we design a topologically reweighting strategy to adaptively weigh the training of edges based on their TE. In addition, to address the limited supervision provided by few minority edges and further enhance the generalizability of our proposed method, following the mixup strategy in~\cite{zhai2022understanding}, we develop a novel topological wedge-based mixup strategy that generates synthetic edges within wedges that are selected based on high TE edges. Finally, we adaptively integrate the two strategies into a novel topological imbalance strategy for edge classification, TopoEdge, to further boost the performance. To demonstrate the effectiveness of our proposed strategies, we first construct a comprehensive imbalance edge classification evaluation benchmark consisting of diverse real-world edge classification datasets that exhibit skewed label distributions. The results reveal TopoEdge is able to significantly improve edge classification performance across various GNN backbones in various settings.

Our main contributions are summarized as follows:
\vspace{-0.75ex}
\begin{itemize}[leftmargin=*]
    \item \textbf{Uncovering a Novel Research Paradigm in Edge Classification:} Through a meticulous exploration into the domain of edge classification, we identified a notable and previously 
    unexplored challenge: topological imbalance. To measure this imbalance with precision, we introduced the innovative Topological Entropy (TE) method, which quantitatively assesses the extent of topological imbalance present within edges' local subgraphs.
    \item \textbf{Algorithmic Design to Address Topological Imbalance:}
    Given the challenges posed by the topological imbalance in edge classification, we pioneered a comprehensive algorithmic solution.
    This first involves a topological reweighting technique, aiming to enhance the training of high TE edges. Complementing this, we devised a unique TE wedge-based mixup approach, which integrates the generation of synthetic training examples derived from high-entropy edges, thereby reinforcing the training efficacy and alleviating the identified imbalance. Ultimately, we combine the above two strategies into a novel Topological imbalance strategy, TopoEdge, for edge classification.
    \item \textbf{SOTA Performance on Real-world Datasets:} To validate the practical implications of our method, we assembled an edge classification benchmark suite with datasets coming from diverse real-world scenarios/applications. Our experiments showcase the ability of the proposed TE-based solutions to help mitigate the imbalance issues in edge classification, and the proposed TopoEdge nearly always surpasses existing methods.

\end{itemize}

\section{Notations and Preliminary }

Let $G = (\mathcal{V}, \mathcal{E}, \mathbf{X}, \mathbf{S})$ be an undirected attributed graph, where $\mathcal{V} = \{v_i\}_{i = 1}^{n}$ is the set of $n$ nodes (i.e., $n = |\mathcal{V}|$) and $\mathcal{E} \subseteq \mathcal{V}\times \mathcal{V}$ is the set of $m$ observed training edges (i.e., $m= |\mathcal{E}|$) with $e_{ij}$ denoting the edge between the node $v_i$ and $v_j$. $\mathbf{X}\in\mathbb{R}^{n\times d}$ represents the node feature matrix with dimension $d$ and $\mathbf{S}\in\mathbb{R}^{m\times d'}$ represents the edge feature matrix with dimension $d'$. The observed $m$ training edges compose the adjacency matrix $\mathbf{A}\in\{0, 1\}^{n\times n}$ with $\mathbf{A}_{ij} = 1$ if an observed edge exists between node $v_i$ and $v_j$ and $\mathbf{A}_{ij} = 0$ otherwise, which serves for the computational graph for message-passing. The diagonal matrix of node degree is notated as $\mathbf{D}\in\mathbb{Z}^{n\times n}$ with the degree of node $v_i$ being $d_i = \mathbf{D}_{ii}=\sum_{j = 1}^{n}\mathbf{A}_{ij}$. The neighborhood node set of node $v_i$ is denoted as $\mathcal{N}_i$. In edge classification, given the edge class label matrix $\mathbf{Y}\in\{0, 1\}^{n\times n\times C}$ with $C$ being the total number of edge classes and denote $\{\mathcal{E}_k\}_{k = 1}^{C}$ as the set of labeled edges in class $k$, we aim to train an edge classifier based on edges in the training set $\mathcal{E}^{\text{Tr}}$ to correctly classify edges in the validation/test set $\mathcal{E}^{\text{Val/Test}}$ across all the classes.

\section{Topological Imbalance and Topological Entropy}
In this section, we first empirically demonstrate that considering the edge imbalance issue merely from the quantitative perspective may result in inferior performance in edge classification. This observation motivates the investigate of edge-level imbalance from the topology perspective. Inspired by prior works that have shown varying local subgraph patterns of nodes cause varying classification performance\cite{mao2023demystifying, zhu2021graph,chen2021topology, song2022tam}, we propose a new type of imbalance in edge classification, topological imbalance, which is defined as the imbalanced local subgraph patterns around each edge that can be quantified by local class distribution variance. Hence we develop a metric, Topological Entropy (TE), to quantify the local class distribution variance, and further verify its relationship with performance discrepancy. Based on the observations, we develop Topological Reweight to emphasize edges with higher TE, and empirical experiments show the effectiveness of our proposed method.

\begin{figure}[t]
     \centering
     \includegraphics[width=0.8\columnwidth]{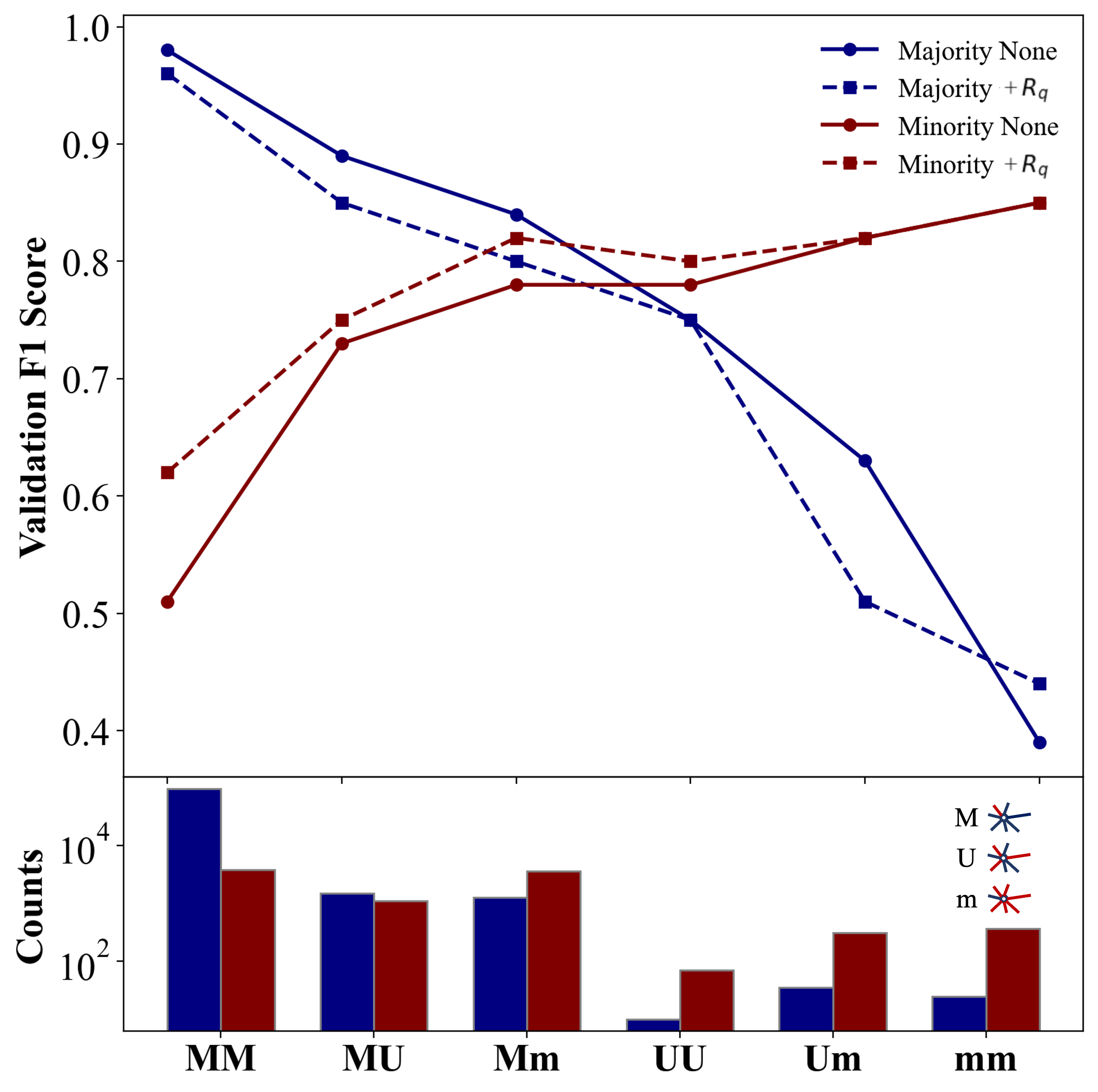}
     \vspace{-1ex}
     \caption{We visualize the validation F1 score for majority and minority class of GCN and GCN+$R_q$ across different edge categories on Epinions dataset. Specifically, edge categories are defined by the two node endpoints being categorized as mostly majority (M), mostly minority (m), or uncertain (U), and we visualize the distribution of edges within the dataset. }
     \label{edge_cat1}
\end{figure}

\subsection{Motivation}~\label{sec-te_moti}
To mitigate the imbalance issue in edge classification, the most direct approach is to implement strategies to mitigate quantity imbalance issues. Hence, we first integrate one of the well-established quantity imbalance techniques, quantity reweight ($R_q$)~\cite{huang2016learning}, with a well-known Graph Neural Network (GNN) architecture, the Graph Convolutional Network (GCN)~\cite{kipf2016semi}, and then conduct edge classification across three datasets. More specifically, for edge $e_{ij} \in \mathcal{E}^{Tr}$ that belongs to class $k$ (i.e., $\mathbf{Y}_{ijk} =1$), the quantity reweight $w_{ij}^{q}$ that is applied to it during the training phase to emphasize the minority class is defined as follows:
\begin{equation}\label{eq-qr}
    w_{ij}^{q} = \frac{|\mathcal{E}^{\text{Tr}}|}{|\{\mathcal{E}_k\}_{k = 1}^{C}|}
\end{equation}
where $|\mathcal{E}^{\text{Tr}}|$ is the number of training edges, and $|\{\mathcal{E}_k\}_{k = 1}^{C}|$ is the number of labeled edges in class $k$. 

As shown in Figure~\ref{fig-gcn_qr} and 
~\ref{fig-gat_qr}, the overall Macro-F1 decreases after equipping quantity reweight strategy to backbones, indicating that this strategy alone which merely addresses the global class distribution is insufficient to achieve consistent performance improvement across all classes. 
This observation compels a thorough examination of the performance disparity among edges through a local lens. To undertake this detailed investigation, as edges are participated by their end nodes, we propose categorizing edges according to the local class distribution (LCD) of their end nodes. 

The categorization process begins with the identification of the edge's end nodes and the computation of the ratio of majority edges within their immediate neighborhoods. Utilizing a predefined threshold, the process categorized each node as mostly majority (M), mostly minority (m), or uncertain (U) based on the corresponding majority ratio. Subsequently, the edge's category is a combination of its end nodes' categories, giving a concise representation of the edge's local class context in the graph. For example, for edge $e_{ij}$, if $v_i$'s category is \textbf{M}, $v_j$ category is \textbf{m}, then edge $e_{ij}$'s category is set as \textbf{Mm}. This nuanced approach effectively captures the edge's local topological context and allows us to analyze the graph's structure in detail, revealing insights into the underlying connectivity patterns. To simplify, the algorithm is designed only for binary classification.

Utilizing the categorization framework delineated earlier, we conduct an in-depth analysis to assess the classification performance across different edge categories, examining both the original GCN model and its extension with $R_q$. As illustrated in Figure~\ref{edge_cat1}, the incorporation of $R_q$ predominantly benefits the minority class, albeit at the expense of diminishing the performance within the majority class for most categories.

More specifically, for the majority class, the most insignificant performance decrease is observed in \textbf{MM}, where the backbone model's performance is nearly perfect, thus applying $R_q$ for edges in this category will negligibly impact overall efficacy. However, in categories characterized by substantial variation in local class distribution (\textbf{MU}, \textbf{Mm}, \textbf{UU}, \textbf{Um}) — with the exception of \textbf{UU}, which comprises a minor quantity of majority edges (counts $\approx 10$) — we observe a pronounced decline in performance. Intriguingly, for category \textbf{mm}, the application of $R_q$ is conducive to performance enhancement. For the minority class, the most significant improvement is gained when the end nodes' categories are both $m$, and then the improvement is decreased across categories \textbf{MU}, \textbf{Mm}, \textbf{UU}, \textbf{Um}, where the end nodes' local distribution varies a lot. In \textbf{mm}, the $R_q$ can no longer improve its performance since the model has reached its latent capacity and performs best.

Based on the above, to strategically design a method on top of $R_q$ to increase the overall edge classification performance, we need to further emphasize the edges with large local class distribution variance. This leads to the next section where we design a novel metric to quantify the level of edges' local class distribution variance.

\vspace{-2ex}
\subsection{Topological Entropy (TE)}
Inspired by the recent work in imbalanced node classification~\cite{chen2021topology, song2022tam}, the local distribution variance can be characterized using the label diffusion method. As label diffusion is based on the receptive field in the computational graph, it is a better approximation of the GNNs message passing scheme, hence the local distribution variance characterized by label diffusion will provide more insight than characterized purely based on label distribution in end nodes' one-hop neighborhood.

For the metric to quantify the local distribution variance, first we obtain label encodings of each node $v_i$ from its incident edges $\{e_{ij}|v_j \in \mathcal{N}_i\}$ by:
\begin{equation}\label{eq-hi}
    \mathbf{h}_{i} = \mathbb{E}_{v_j \sim \mathcal{N}_i}{\mathbf{Y}_{ij}},
\end{equation}
where $\mathbf{h}_{i}\in\mathbb{R}^{C}$ denotes the probability distribution of the node $v_i$'s association to the edge classes according to their incident edges. To consider higher-order neighborhood impact, we obtain the $L^{\text{th}}$-layer label encodings $\mathbf{H}^{L}$ by message-passing:
\begin{equation}
    \mathbf{H}^{L} = \widetilde{\mathbf{A}}\mathbf{H}^{L - 1},
\end{equation}
where $\widetilde{\mathbf{A}} = \mathbf{D}^{-1}\mathbf{A}$ is the normalized adjacency matrix and specifically, $\mathbf{H}^{0} = \mathbf{H}$, coming from Eq.~\eqref{eq-hi}.  
Since we have the label distribution of $L$-hop subgraphs of the ending nodes $v_i, v_j$ of each edge $e_{ij}$, we then can directly average $\mathbf{H}^{L}_i, \mathbf{H}^{L}_j$ to obtain the label distribution $\mathbf{G}_{ij}^{L} \in \mathbb{R}^{C}$ of the local subgraph around that edge $e_{ij}$.
\begin{equation}
    \mathbf{G}_{ij}^L = \frac{1}{2}(\mathbf{h}^L_i + \mathbf{h}^L_j),
\end{equation}
where $\mathbf{G}_{ijk}^{L}$ indicates the proportion of the $L$-hop local subgraph around the edge $e_{ij}$ belonging to the class $k$. In this way, if the end nodes of an edge have identical and certain class distributions (i.e., \textbf{MM} and \textbf{mm}), the distribution will be highly skewed, and the uncertainty will be small. For the edge with large class distribution variance (i.e., edges in \textbf{MU}, \textbf{Mm}, \textbf{UU}, and \textbf{Um}), the distribution will be more uniform, and the uncertainty will be large.

Hence, after we have a well-established metric $\mathbf{G}_{ij}^{L}$ to quantify the local topology pattern around each edge, we can further quantify its entropy to measure its class uncertainty, defined as: 
\begin{definition}
\textbf{Topological Entropy (TE):} The uncertainty of the class distribution of the $L^{\text{th}}$-layer local subgraph around each edge $e_{ij}$ is calculated as:
\begin{equation}\label{eq-TE}
    \text{TE}_{ij}^{L} = - \sum_{k=1}^{C}\textbf{G}_{ijk}^{L}\log(\textbf{G}_{ijk}^{L})
\end{equation}
\end{definition}
In this way, the edges with large class distribution variance would have a higher Topological Entropy (TE) value.

\begin{figure}
\vspace{2ex}
     \centering
     \includegraphics[width=.91\columnwidth]{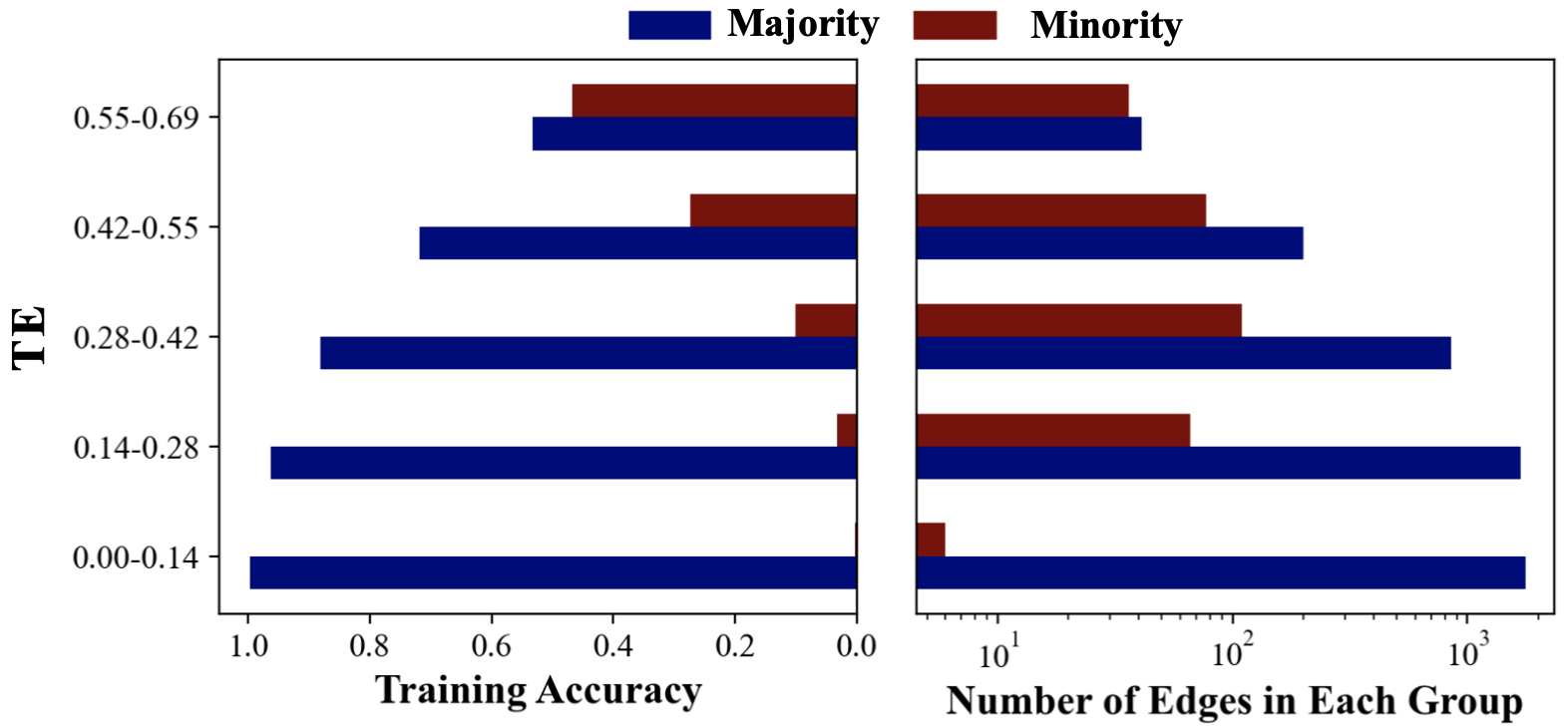}
     \vspace{-1.25ex}
     \caption{We visualize the empirical discrepancy (in terms of training accuracy) between majority and minority edges when grouped by their TE (along with the number of edges in each group) and using GCN on the Epinions dataset.}
     \label{topo-tracc}
     \vspace{-1.75ex}
\end{figure}

To demonstrate the accuracy of Topological Entropy (TE) in reflecting local class distribution variance, we undertook an empirical study on the Epinions dataset. The experiment aimed to confirm that edges within categories marked by low local distribution variance—specifically \textbf{MM} and \textbf{mm} from Figure~\ref{edge_cat1}—would exhibit lower TE values. As shown in Figure~\ref{topo-tracc}, majority edges are easier to train correctly than minority edges, which aligns with the patterns in Figure~\ref{edge_cat1}. Figure~\ref{topo-tracc} also reveals that majority edges with lower TE values (i.e., TE $\le 0.28$) are indeed easy to train correctly, affirming the relatively high validation F1 scores for majority edges in \textbf{MM} and \textbf{mm} as observed in Figure~\ref{edge_cat1}. Conversely, minority edges with low TE struggled in training, mirroring their lower validation F1 scores in Figure~\ref{edge_cat1}. Additionally, the distribution trends of majority and minority edges across varying TE values further corresponded with the patterns observed (the distribution of majority edges skewed towards lower TE values, and the distribution of minority edges is rather uniform) in Figure~\ref{edge_cat1}, confirming the effectiveness of TE in capturing the anticipated variance. These findings affirm the methodological soundness of TE, laying a solid foundation for subsequent methodological development.

\section{Topological Reweight}\label{sec-top-tr}
Building upon Section~\ref{sec-te_moti}, we expect to enhance the training weights of labeled edges that have larger TE values, thereby enabling those edges that are more representative of their corresponding classes and hence lead to better performance for unlabeled edges.

For any edge $e_{ij} \in \mathcal{E}^{Tr}$, we design a TE reweight method to assign edge weight $w^{t}_{ij}$ as:
\begin{equation}\label{trw}
    w^{t, L}_{ij} = exp(\frac{\text{TE}^L_{ij}}{t}),
\end{equation}
where $t$ is the temperature controlling the sharpness of the weight distribution and as $t \rightarrow 0$, the weight distribution becomes more uniform and hence it approaches the unweighted version. Then the training loss $L$ for edge classification equipped with TE reweight is:
\begin{equation}\label{eq-lossTE}
    \mathcal{L}^{\text{TE}} = \frac{1}{|\mathcal{E}^{Tr}|}\sum_{e_{ij} \in \mathcal{E}^{Tr}}{w_{ij}^{t, L}\mathcal{L}^{\text{CE}}(\mathbf{E}_{ij}, \mathbf{Y}_{ij})}
\end{equation}
where $\mathbf{E}_{ij}$ is the embedding obtained from any GML model (e.g., a GNN)
followed by a classifier and having $\mathcal{L}^{\text{CE}}$ denote the cross entropy loss. This is then the loss function of TE reweight ($R_t$).

\begin{figure}
     \centering
     \begin{subfigure}[b]{0.225\textwidth}
         \centering
         \includegraphics[width=\textwidth]{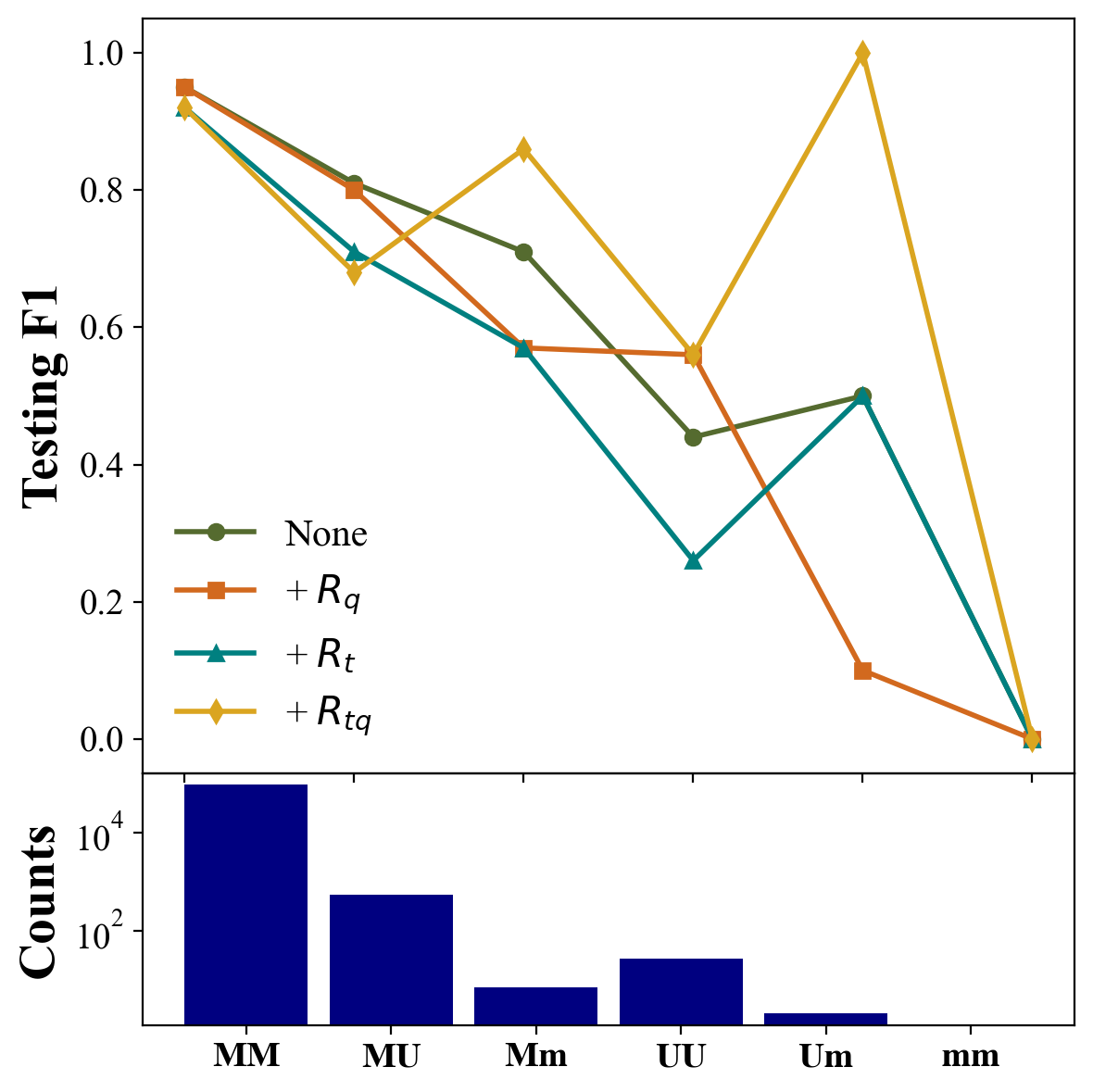}
         \caption{}
         \label{fig-maj_cat}
     \end{subfigure}
     \begin{subfigure}[b]{0.225\textwidth}
         \centering
         \includegraphics[width=\textwidth]{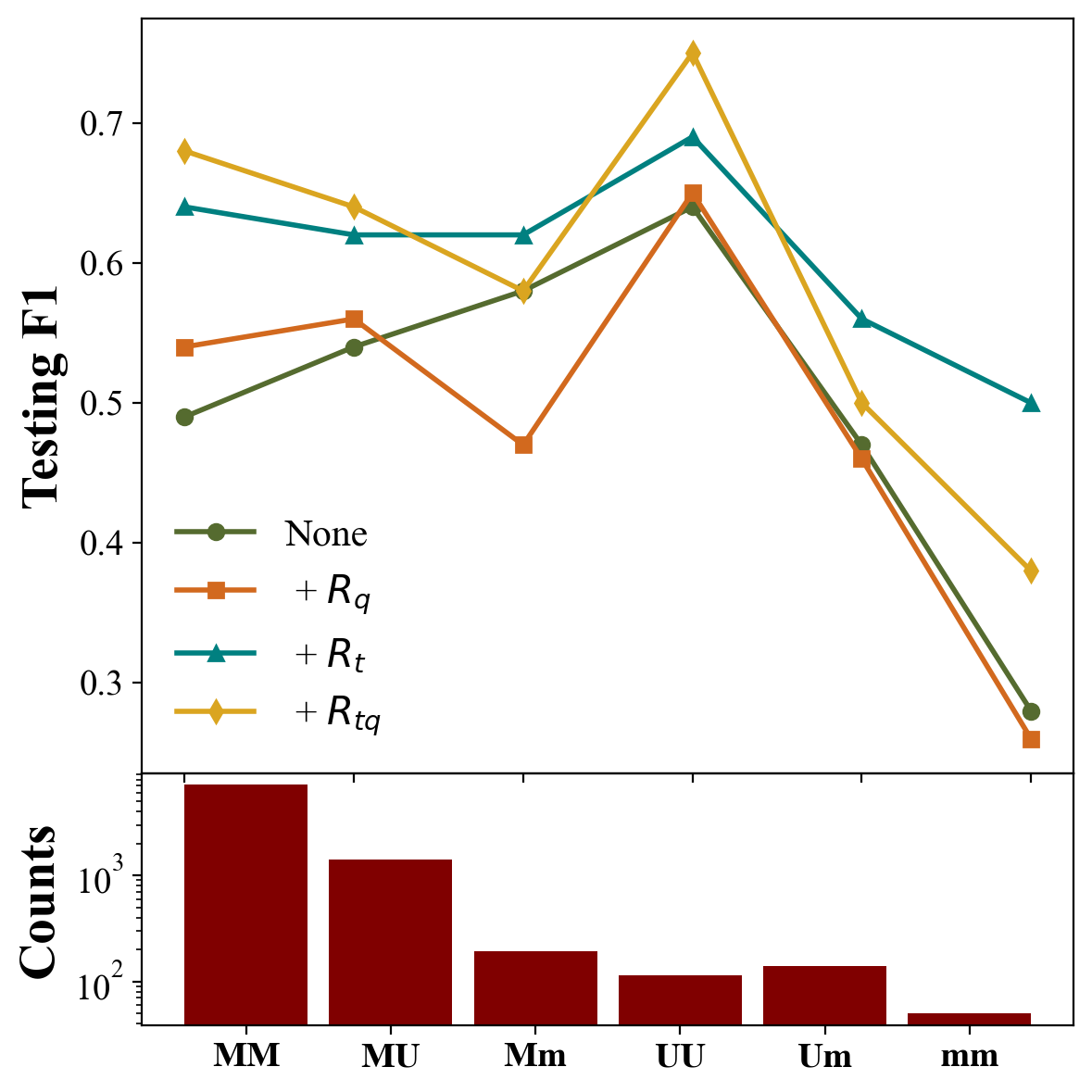}
         \caption{}
         \label{fig-min_cat}
     \end{subfigure}
     \vskip -1.75ex
     \caption{After applying $R_{tq}$ which strategically emphasizes edges with high local distribution variance on top of $R_q$, significant performance improvement can be observed for majority class \textbf{(a)} and minority class \textbf{(b)}, with the most significant improvement can be observed for edges in relatively large class distribution variance categories (i.e., edges in \textbf{MU}, \textbf{Mm}, \textbf{UU}, and \textbf{Um} categories).
     }
     \label{fig-intro-qr}
     \vskip -1.75ex
\end{figure}

Based on the insight in Section~\ref{sec-te_moti}, we adaptively integrate TE reweight with the previously introduced quantity reweight method into a novel Topological Reweight ($R_{tq}$) strategy to include both the global class distribution and local class distribution variance into consideration. The final weight can be computed as:
\begin{equation}\label{eq-fw}
    w_{ij} = \theta w_{ij}^{t, L} + (1-\theta) w_{ij}^{q}
\end{equation}
where $\theta$ is a hyperparameter that controls the trade-off in the contribution between topological reweight and quantitative reweight. Correspondingly the training loss becomes:
\begin{equation}\label{lt}
    \mathcal{L}^{\text{T}} =\frac{1}{|\mathcal{E}^{Tr}|}\sum_{e_{ij} \in \mathcal{E}^{Tr}}w_{ij}\mathcal{L}^{\text{CE}}(\mathbf{E}_{ij}, \mathbf{Y}_{ij})
\end{equation}

To validate the effectiveness of the proposed methods, we empirically compare the edge classification performance of quantitative reweight ($R_q$), TE reweight ($R_t$), and their combined version Topological reweight ($R_{tq}$). First, we visualize the classification performance for majority and minority edges across different categories in Epinions dataset. As shown in Figure~\ref{fig-maj_cat} and Figure~\ref{fig-min_cat}, $R_{tq}$ can indeed improve the performance for edges in \textbf{MU}, \textbf{Mm}, \textbf{UU}, and \textbf{Um}, which aligns with our intuition to improve performance on edges with high local distribution variance. 

To empirically measure the classification performance of different methods across all the classes, we follow other imbalance classification work~\cite{chen2021topology,liu2023topological, sun2022position, ma2023class} and adopt balanced accuracy (average accuracy across different classes) and Macro-F1 (average F1 score across different classes) as the performance metric. As shown in Table~\ref{tab-topo-result}, both $R_t$ and $R_{tq}$ consistently yield enhancements in terms of balanced accuracy and Macro-F1 scores. This outcome empirically substantiates the efficacy of the proposed reweight method against $R_q$ where only performance improvement in minority classes can be observed. The fact that applying topology reweight that addresses the local structure disparity performs better than quantity reweight suggests that local class distribution is more important for performance improvement, which is reasonable because in GNN message passing nodes and edges only receive messages from the local subgraph. Also, combining quantity reweight with topology reweight adaptively can take advantage of the information in both global class distribution and local class distribution and mitigate both quantity and topological imbalance issues, hence it depicts the optimal performance.

\section{Mixup based on TE}\label{rw}
In Section~\ref{sec-top-tr}, we present the Topological Reweight ($R_{tq}$) mechanism to emphasize the edges according to their TEs.  However, merely incorporating topological reweight may not address the limited supervision provided by few minority edges, and thus there is no guaranteed performance improvement~\cite{zhai2022understanding, byrd2019effect, sagawa2019distributionally, gulrajani2020search, koh2021wilds}. Moreover, reweighting methods can lead to overfitting~\cite{sagawa2019distributionally}, with test performance potentially degrading over numerous epochs without intervention. One method that may address these issues and further enhance the generalizability of our proposed method is mixup, as it generates synthetic samples to change the original training distribution and provides sufficient supervision. Currently, there are some mixup methods in GML tasks ~\cite{wang2021mixup,kim2023s, han2022g, ma2024fused, li2022graph}, but none of them are designed to mitigate topological imbalance issues in edge classification. Therefore, to improve the generalizability and effectiveness of our proposed method in topological imbalance edge classification tasks, we designed a TE wedge-based mixup method, where we selected the edges with high entropy and its high entropy neighborhood edges to form a ``high TE wedge'', then we generate new sample node by mixup between end nodes of wedges. The overview of the TE wedge-based mixup is shown in Figure~\ref{plt_mixup}.

\begin{table}[t]
\scriptsize
 \vspace{-2.0ex}
\caption{Balanced accuracy (B\_acc) and Macro-F1 (Macro\_F1) across three datasets using the GCN backbone model showing performance of $R_q$, $R_t$, and $R_{tq}$. }\label{tab-topo-result}
\begin{tabular}{cccccc}
\hline
Dataset & Metric & GCN & GCN$+{R_q}$ & GCN$+{R_t}$ & GCN$+{R_{tq}}$ \\
\hline
\multirow{2}{*}{Bitcoin-alpha} & B\_acc & 0.797 $\pm$ 0.005 & 0.857 $\pm$ 0.020 & 0.818 $\pm$ 0.004 & 0.834 $\pm$ 0.008  \\
 & Macro\_F1 & 0.840 $\pm$ 0.005 & 0.790 $\pm$ 0.005 & 0.846 $\pm$ 0.007 & 0.852 $\pm$ 0.003 \\
\hline
\multirow{2}{*}{Epinions} & B\_acc & 0.731 $\pm$ 0.008 & 0.768 $\pm$ 0.007 & 0.756 $\pm$ 0.008 & 0.769 $\pm$ 0.001 \\
 & Macro\_F1 & 0.745 $\pm$ 0.002 & 0.720 $\pm$ 0.001 & 0.756 $\pm$ 0.000 & 0.753 $\pm$ 0.001 \\
\hline
\multirow{2}{*}{Reddit} & B\_acc & 0.715 $\pm$ 0.003 & 0.779 $\pm$ 0.012 & 0.727 $\pm$ 0.035 & 0.762 $\pm$ 0.006 \\
 & Macro\_F1 & 0.733 $\pm$ 0.000 & 0.707 $\pm$ 0.002 & 0.737 $\pm$ 0.007 & 0.740 $\pm$ 0.000\\
\hline
\end{tabular}
\vskip -2.0ex
\end{table}

\subsection{Motivation}
To circumvent the potential pitfalls of the reweighting method that we stated above, and to accentuate edges with elevated entropy, our strategy mandates refinement beyond mere Topological reweight. A corpus of discussions, encapsulated in~\cite{wiles2021fine, sagawa2021extending, zhai2022understanding}, advocate for the incorporation of additional training samples via data augmentation as a viable enhancement to topological reweighting. Wang et al.\cite{wang2021importance} further contend that modifications to the loss function can potentially amplify the efficacy of our proposed approach. Given these considerations, we posit:

\begin{figure}[t]
     \centering
 \includegraphics[width=0.45\textwidth]{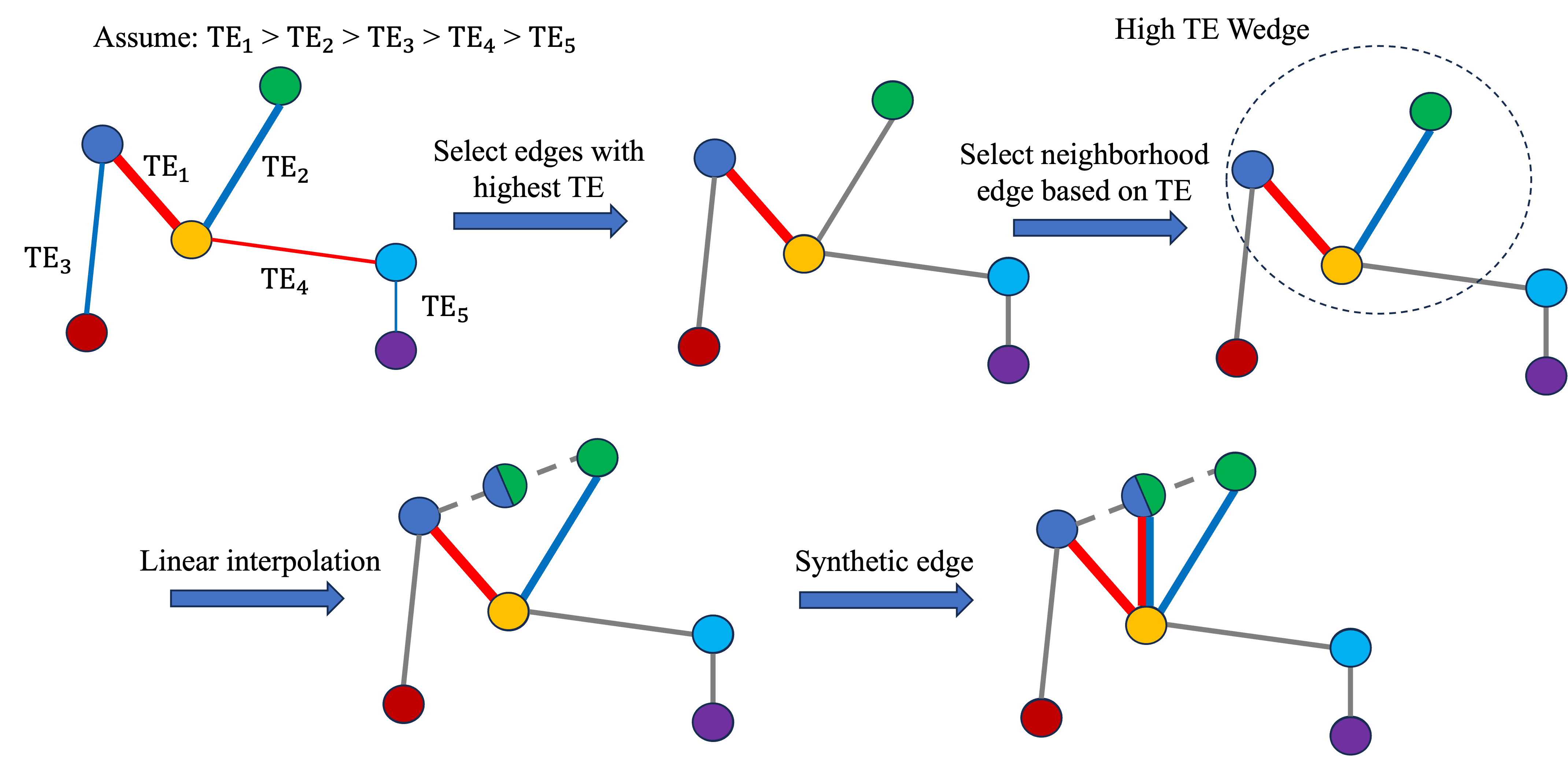}
 \vspace{-2.25ex}
 \caption{Overview of the Topological Wedge-based mixup}\label{plt_mixup}
 \vskip -2.25ex
\end{figure}

\textit{Can we develop strategies that combine these elements and incorporate our understanding of topological imbalance in edge classification?} 

Our exploration led us to the random mixup technique.
At its core, random mixup formulate synthetic samples via linear interpolation between two randomly selected distinct samples. The central tenet postulates that linear interpolations between feature vectors should logically precipitate linear interpolations of their respective targets~\cite{zhang2017mixup}. However, direct adoptions of this technique between any two random samples present a problem for edge classification since random edge interpolations lack semantic coherence~\cite{zhang2017mixup}, and one way to solve that is to perform mixup in wedges. However, randomly selecting wedges to perform mixup cannot address the observations made in the previous section, as we want to emphasize the high TE edges while preserving semantic coherence. Therefore, we extend our wedge-based mixup to develop a novel Topological wedge-based mixup and formally present it in the next section.

\subsection{TE Wedge-based Mixup}
In this section, we delineate the underlying intuition and technical intricacies of the wedge-based mixup we propose. We commence by elucidating the rationale and driving forces behind the adoption of the topological wedge-based mixup in edge classification. Subsequently, the mechanics of this approach are elaborated upon. Our methodology hinges on two primary modules. The \emph{High TE Wedge Selection Module} is employed to curate high TE wedges. Within these selected wedges, a \emph{Wedge-mixup Module} facilitates node mixup, engendering synthetic nodes and edges and integrating these edges with topological weight, culminating in the final loss function. This evolved loss function supplants the original variant presented in Eq.~\eqref{lt}, paving the way for the training of the edge classifier. Below we present the details of each module.

\subsubsection{\textbf{High TE Wedge Selection Module}:} We first define the concepts and notions about wedge: 
\begin{definition}
\vspace{-1ex}
A wedge $ \mathcal{W}_{i,c,j}$ is a set comprising two distinct undirected edges $e_{ci}$ and $e_{cj}$ intersecting at a singular shared node $v_c$, which is referred to as the ``centric node'', the node $i$ and node $j$ are called ``end nodes:''
\begin{equation}\label{w}
\small
    \mathcal{W}_{i,c,j} = \{e_{ci}, e_{cj} | e_{ci}, e_{cj} \in \mathcal{E}, v_i,v_c,v_j \in \mathcal{V}, i \neq j\}
\end{equation}
\end{definition}
To address the observation we made in the previous section and emphasize edges with high TE, we begin by determining the TE values for edges within the training set. Then we select $K$ number of training edges based on their TE values. Specifically, edges with the highest TE are most likely to be selected. This methodology gives rise to a subset of selected edges, denoted as \( \mathcal{E}_{1} \subseteq \mathcal{E}_{Tr} \), and its associated edge label set \( \mathbf{Y}_{1} \). For each edge \( e_{ci} \) within \( \mathcal{E}_{1} \), we discern an edge \( e_{cj} \) that is incident with it. In cases where multiple incident edges are present, preference is accorded to those with elevated TE values. Consequently, we identify the incident edge set \( \mathcal{E}_{2} \subseteq \mathcal{E}_{Tr} \), its pertinent edge labels \( \mathbf{Y}_{2} \), and high TE wedge set \( \mathcal{W} \).

\subsubsection{\textbf{Wedge-Mixup Module}:} 

With the aid of the GNN encoder, node embeddings, denoted as \( Z \), are procured. Then for every wedge $\mathcal{W}_{i,c,j}$, we can generate a new synthetic node $v_s$ between node $v_i$ and $v_j$ with embedding as:
\begin{equation}\label{node}
\small
    \mathbf{Z}_{v_s} = \lambda \mathbf{Z}_{v_i} + (1 -\lambda) \mathbf{Z}_{v_j}
\end{equation}
where $\lambda \sim Beta (\alpha, \alpha)$, and $\alpha$ is a hyperparameter that controls the Beta distribution. In this paper, we set $\alpha$ to be $4.0$.
Then we can get the new synthetic edge $e_{cs}$'s edge embedding as 
$\mathbf{E}^{'}_{e_{cs}} = \mathbf{Z}_{v_c} \Vert \mathbf{Z}_{v_s}$ 
and the synthetic edge feature embedding can be generated as 
$\mathbf{S}_{e_{cs}} = \lambda \mathbf{S}_{e_{ci}} + (1-\lambda) \mathbf{S}_{e_{cj}}$. 
Then the final embedding for the synthetic edge is defined as 
$\mathbf{E}_{e_{cs}} = \mathbf{E}^{'}_{e_{cs}} \Vert \mathbf{S}_{e_{cs}}$. 
Let the synthetic edge set to be $\mathcal{E}_3$. At this point, we can send $\mathbf{E}_{\mathcal{E}_3}$ into MLP classifier $f$ to get the outcome $f({\mathbf{E}_{\mathcal{E}_3}})$ that can be used for the following-up module. However, directly using this in practice will bring trouble as the golden label can not be linearly interpolated. Since the objective of the TE wedge-based mixup loss is to minimize the differences between synthetic edge prediction outcome and synthetic edge label, and the synthetic edges are generated through linear interpolation between edges in $\mathcal{E}_1$ and $\mathcal{E}_2$. 
For implementation, as suggested in ~\cite{zhang2017mixup}, the loss can be computed by linearly combining the loss of edges in $\mathbf{Y}_1$ and $\mathbf{Y}_2$. Thus, the loss function for our TE wedge-based mixup is:
\begin{equation}\label{lx}
    \mathcal{L}^{\text{X}} = \lambda \mathcal{L}^{\text{CE}}(f(\mathbf{E}_{\mathcal{E}_3}), \mathbf{Y}_1) + (1-\lambda) \mathcal{L}^{\text{CE}}(f(\mathbf{E}_{\mathcal{E}_3}, \mathbf{Y}_2) 
\end{equation}
where $\mathcal{L}^{\text{CE}}$ is the cross entropy loss. Since incorporating a linear combination of Eq.~\eqref{lx} with original loss into the final loss function deviates from the traditional ERM paradigm, which is congruent with our earlier discourse and is anticipated to bolster the efficiency of the methodology, we can get the final loss function as:
\begin{equation}\label{l}
    \mathcal{L}= \mathcal{L}^{\text{Original}} + h\mathcal{L}^{\text{X}} 
\end{equation}
where $h$ balances the TE wedge-based mixup and original losses.

\begin{table}[t]
\scriptsize
\setlength\tabcolsep{1pt}
\caption{Classification performance of backbones before and after applying random edge-based mixup ($X_e$), TE edge-based mixup ($X_{te}$), random wedge-based mixup ($X_w$), and TE wedge-based mixup ($X_{tw}$) strategies. }
\label{mixup}
\begin{tabular}{c|cc|cc|cc}
\hline
Dataset & \multicolumn{2}{c|}{Bitcoin-alpha} & \multicolumn{2}{c|}{Epinions} & \multicolumn{2}{c}{Reddit} \\ \hline
Metric  & B\_acc          & Macro\_F1         & B\_acc         & Macro\_F1        & B\_acc            & Macro\_F1          \\ \hline
GCN                 & 0.797 $\pm$ 0.005       &   0.840 $\pm$ 0.005   & 0.731 $\pm$ 0.008  &  0.745 $\pm$ 0.002 & 0.715 $\pm$ 0.003 & 0.733 $\pm$ 0.000 \\
GCN+$X_e$               & 0.813 $\pm$ 0.004 & 0.851 $\pm$ 0.001   & 0.753 $\pm$ 0.008 &  0.744 $\pm$ 0.001  & 0.719 $\pm$ 0.006 & 0.732 $\pm$ 0.001  \\
GCN+$X_{te}$               & 0.818 $\pm$ 0.005 & 0.851 $\pm$ 0.006   & 0.744 $\pm$ 0.004 &  0.748 $\pm$ 0.002  & 0.720 $\pm$ 0.016 & 0.732 $\pm$ 0.004  \\
GCN+$X_w$               & 0.811 $\pm$ 0.005 & 0.850 $\pm$ 0.002   & 0.751 $\pm$ 0.004  &  0.742 $\pm$ 0.002 & 0.710 $\pm$ 0.011 & 0.730 $\pm$ 0.005\\
GCN+$X_{tw}$               & 0.828 $\pm$ 0.010  & 0.852 $\pm$ 0.008   & 0.754 $\pm$ 0.007 &  0.749 $\pm$ 0.002  & 0.728 $\pm$ 0.016 & 0.737 $\pm$ 0.002  \\ \hline
\end{tabular}
\vskip -3.5ex
\end{table}

\begin{figure*}[ht!]
     \centering
 \includegraphics[width=0.9\textwidth]{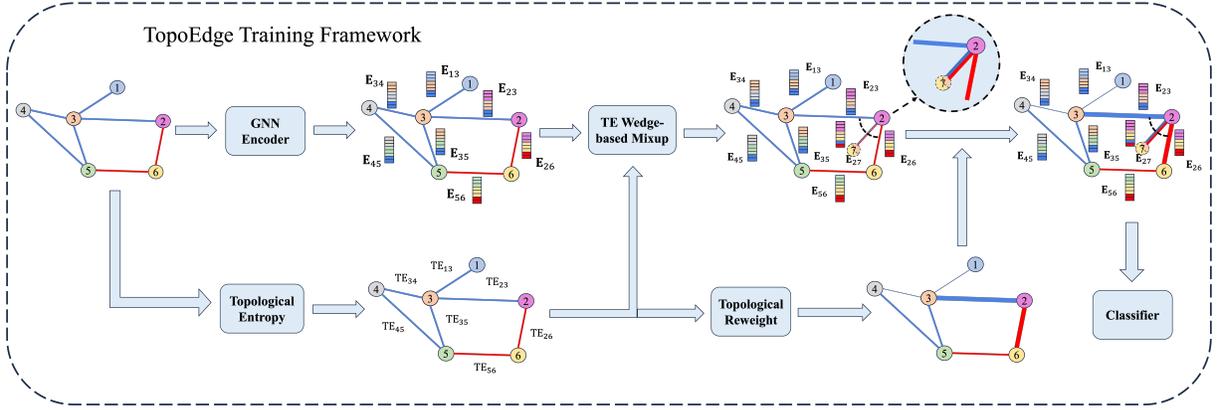}
 \caption{The overall framework of TopoEdge. The training process begins with utilizing the original graph to compute both edge embedding $\mathbf{E}$ and edge TE values. These components facilitate the execution of a TE wedge-based mixup strategy, leading to the creation of synthetic edges and their respective embeddings (i.e., $\mathbf{E}_{27}$ in the figure). Concurrently, TE values inform a topological reweighting process, producing edge weights. The ensemble of edge weights and embeddings forms the input for the final classifier, which will generate output as the edge classification results.}\label{plt_flow}
\end{figure*}

\subsection{Empirical Analysis}
In this section, we provide an empirical analysis of the effectiveness of TE wedge-based mixups ($X_{tw}$), which do not incorporate a topological reweight term in the final loss function, enabling a fair comparison. We contrast this method with random edge-based mixups ($X_e$), TE edge-based mixups ($X_{te}$) where edges with the highest TE scores are preferentially selected for mixups, and standard wedge-based mixups ($X_w$).

The results presented in Table~\ref{mixup} underscore the superior performance of our TE wedge-based mixup strategy, which consistently outperforms alternative methods in both balanced accuracy and Macro-F1. This confirms the hypothesis that prioritizing high TE edges can effectively address the issue of topological imbalance.

Moreover, the observation that TE edge-based mixups surpass random edge-based mixups, yet fall short of TE wedge-based mixups, reinforces the value of our approach. While emphasizing high TE edges alone does yield performance improvements, the integration of these edges with a wedge-based strategy — which also tackles semantic coherence — proves to be more advantageous. This affirms the validity of our method and aligns with our initial objectives, demonstrating the effectiveness of combining high TE edge prioritization with random wedge-based mixup techniques.

\section{Overall TopoEdge Framework}

To further boost the performance, we can adaptively integrate the proposed topological reweight with TE wedge-based mixup as:
\begin{equation}\label{ltx}
    \mathcal{L}^{\text{tX}} = \lambda w_{\mathcal{E}_3} \mathcal{L}^{\text{CE}}(f(\mathbf{E}_{\mathcal{E}_3}), \mathbf{Y}_1) + (1-\lambda) w_{\mathcal{E}_3}\mathcal{L}^{\text{CE}}(f(\mathbf{E}_{\mathcal{E}_3}, \mathbf{Y}_2) 
\end{equation}
Note that the weight for the synthetic edges will be the linear integration between the weights of TE wedges' end edges. Then similarly we can linearly combine Eq.~\eqref{ltx} with Eq.~\eqref{lt} to get the final loss function as:
\begin{equation}\label{l}
    \mathcal{L}= \mathcal{L}^{\text{T}} + h\mathcal{L}^{\text{tX}} 
\end{equation}
Where $h$ is a hyperparameter that scales the topological reweight loss and topological wedge-based mixup loss. This is our final topological imbalance strategy for edge classification: \textbf{TopoEdge}. The pseudo-code of TopoEdge is shown in Algorithm~\ref{alg2}, and the overall framework of TopoEdge is shown in Figure~\ref{plt_flow}.

\begin{algorithm}[]
\caption{Overall TopoEdge Framework}
\small 
\label{alg2}
\begin{algorithmic}[1]
    \State \textbf{Input} Edge embedding $\mathbf{E}$, training edges $\mathcal{E}^{Tr}$, training edge labels $\mathbf{Y}$, \newline classifier $f$,  hyperparameters $K, \alpha, t, \theta, h$
    \State Get the quantity weight from Eq. ~\eqref{eq-qr}
    \State Get the TE weights for edges in $\mathcal{E}^{Tr}$ from Eq. ~\eqref{eq-TE} and ~\eqref{trw} 
    \State Get the topological weight from Eq. ~\eqref{eq-fw}
    \State Get the topological training loss $\mathcal{L}^{\text{T}}$ from Eq.~\eqref{lt}
    \State $\lambda \sim Beta (\alpha, \alpha)$
    \State High TE Wedge Selection Module: \begin{itemize}
        \item Select $K$ edges based on TE value to get $\mathcal{E}_{1}$, $\mathbf{Y}_1$
        \item Get incident edge set $\mathcal{E}_{2}$, $\mathbf{Y}_2$, based on TE and $\mathcal{E}_{1}$, \newline to form high TE wedge set $\mathcal{W}$
    \end{itemize}
    \State Wedge-Mixup Module:\begin{itemize}
        \item Get synthetic loss $\mathcal{L}^{\text{tX}}$ from Eq. ~\eqref{ltx} 
    \end{itemize}
    \State Wedge-loss Module: \begin{itemize}
        \item Get final loss $\mathcal{L}$ from Eq. ~\eqref{l} 
    \end{itemize}
    \State \Return $L$
\end{algorithmic}

\end{algorithm}

\vspace{-1ex}
\section{Experiments}\label{sec-experiments}
In this section, we provide a comprehensive study of the effectiveness of our methods on real-world edge classification tasks. 

\subsection{Datasets} 
In this section, we present the experimental datasets employed in this study. As highlighted earlier, it is evident that issues related to topology imbalance can be observed in edge classification across various applications. Thus, we collect and formulate six benchmark imbalanced real-world graph datasets with relevant features. Below are brief descriptions of each dataset, and the basic statistics of the datasets are provided in Table~\ref{dataset}.
\begin{itemize}[leftmargin=*]
    \item Bitcoin-alpha~\cite{derr2017signed} is a cryptocurrency transaction trust network, where a node represents the user and a directed edge represents ratings (how much one user trusts/distrusts another user). The edge features are text embeddings of user comments.  
    Note that ratings ranging -10 (total distrust) to +10 (total trust) can be discretized into different edge types. In binary classification, trust (ratings 1 to 10) is the majority class ($93.8\%$), while distrust (ratings -10 to -1) is the minority ($6.3\%$). For multi-class classification, weak trust ($90.3\%$), strong trust ($3.5\%$), weak distrust ($3.3\%$), and strong distrust ($3.0\%$) are considered. 
    \item HSPPI is a collection of homo sapiens protein-protein interaction networks based on \href{https://string-db.org/cgi/download?sessionId=bzZ3BKKHe9Tc&species_text=Homo+sapiens}{String database}~\cite{szklarczyk2023string}. Nodes represent proteins, and node features are protein sequences (acquired from \href{https://www.ebi.ac.uk/proteins/api/doc/}{EBI protein API}~\cite{nightingale2017proteins}) that reflect the protein amino acids sequence information. The two edge types are indirect functional interactions ($76.3\%$) and direct physical interactions ($23.7\%$).

\begin{table}[t]
\small 
\setlength\tabcolsep{2.25pt}
\vspace{-1ex}
\caption{Basic statistics of the imbalanced benchmark datasets. The 4 largest frequency ratios in each dataset are represented as $L_0$ through $L_3$. }

\centering
\vspace{-2ex}
\begin{center}
\begin{tabular}{c|cccccc|c}
\hline
Datasets                   & \# Nodes     & \# Edges     & $L_0$     & $L_1$     & $L_2$     & $L_3$     & \# Labels \\ \hline
Bitcoin-alpha              & 3,784       & 24,207      & 90.3\% & 3.5\% & 3.3\% & 3.0\%    & 2 or 4            \\

HSPPI     & 17,895     &  1,008,006      & 76.1\% & 23.9\% & - & - &   2    \\

NID                 & 109,802      & 431,372     & 62.0\% & 14.7\% & 14.3\% & 3.0\%  &  9          \\
Epinions                      & 81,350     & 763,538      & 85.4\% & 14.6\%  & -      & -          &  2            \\ 
Reddit                     & 67,180     & 901,056      & 90.5\% & 9.5\%  & -      & -          &  2 \\
MAG                     & 40,000     & 141,725      & 58.6\% & 14.2\%  & 13.7\%      & 7.3\%           &  10 \\ \hline
\end{tabular}
\end{center}
\vskip -2ex
\label{dataset}
\end{table}

    \item NID is formulated from NetFlow-based IoT network dataset \href{https://staff.itee.uq.edu.au/marius/NIDS_datasets}{NF-ToN-IoT}~\cite{sarhan2021netflow}, where node represents user's IP address, and edge represents interactions between users. There are $9$ classes in total, and $62.0 \%$ of the interactions are injection, $14.7\%$ are DDos, $14.3\%$ are benign (normal unmalicious flows), and $3.2\%$ are scanning. Edge features are network flow statistics (flow duration time, flow bytes, packets, etc) between users.

    \item Epinions~\cite{guha2004propagation} is a product review website where nodes represent each user and directed edges represent the trust/ distrust relationship from one user to another. Each user has reviews for different products. During the data filtering processing, we first remove the users who never give reviews, and then we concatenate the text embeddings of the source node with the target node to form the edge feature. We eventually have $85.4\%$ trust and $14.6\%$ distrust edges.

    \item Reddit~\cite{kumar2018community} is a hyperlink network 
    with communities as nodes and posts between them as edges. The edge labels describe the sentiments of the posts in source communities towards the posts in target communities, where $90.5\%$ are neutral or positive, and $9.5\%$ sentiments are negative. The edge features are text property vectors between the communities.

    \item MAG~\cite{sinha2015overview} is a citation network where nodes and edges represent the scholars and papers coauthored by scholars, respectively. Edge labels correspond to the field of study for the paper. Edge features are bag-of-words of paper abstracts. In this paper, we adopt the setting in ~\cite{wang2023efficient}, and the majority class ratio is $58.6\%$.
    
\end{itemize}

\begin{table*}[t]
\footnotesize
\setlength\tabcolsep{4pt}
\caption{Binary classification results on four real-world datasets indicate that our proposed method can achieve steady improvement in terms of both balanced accuracy and Macro-F1 compared with backbone GNNs, and applying TopoEdge can achieve comparable performance as SOTA baselines.}\label{fig-binary}

\vskip -2ex
\begin{tabular}{|c|cc|cc|cc|cc|}
\cline{2-9}
 \multicolumn{1}{c|}{} &
  \multicolumn{2}{c|}{Bitcoin-alpha} &
  \multicolumn{2}{c|}{HSPPI} &
  \multicolumn{2}{c|}{Epinions} &
  \multicolumn{2}{c|}{Reddit} \\ \cline{2-9}
 \multicolumn{1}{c|}{} & B\_acc & Macro\_F1 & B\_acc & Macro\_F1 & B\_acc & Macro\_F1 & B\_acc & Macro\_F1  \\ \hline
ExtWF        & 0.500 $\pm$ 0.000 & 0.484 $\pm$ 0.000 & 0.497 $\pm$ 0.000  & 0.432 $\pm$ 0.000 & 0.500 $\pm$ 0.000 & 0.476 $\pm$ 0.000 & 0.552 $\pm$ 0.000 & 0.570 $\pm$ 0.000 \\ \hline
TER+AER (Geometric)                & 0.871 $\pm$ 0.013 &  0.870 $\pm$ 0.007 & 0.598 $\pm$ 0.043 & 0.514 $\pm$ 0.063  & 0.764 $\pm$ 0.003 &  0.783 $\pm$ 0.002  & 0.789 $\pm$ 0.002 & 0.778 $\pm$ 0.002 \\
TER+AER (Possion)           & 0.868 $\pm$ 0.015 & 0.869 $\pm$ 0.004 & 0.603 $\pm$ 0.037 & 0.520 $\pm$ 0.051  & 0.768 $\pm$ 0.003 & 0.783 $\pm$ 0.002 & 0.786 $\pm$ 0.006 & 0.773 $\pm$ 0.002 \\ \hline
GCN w/o TopoEdge              & 0.797 $\pm$ 0.005       &   0.840 $\pm$ 0.005       & 0.655 $\pm$ 0.003  &  0.667 $\pm$ 0.001        & 0.731 $\pm$ 0.008  &  0.745 $\pm$ 0.002      & 0.715 $\pm$ 0.003 & 0.733 $\pm$ 0.000\\
GCN w/ TopoEdge           & 0.845 $\pm$ 0.004       &   0.858 $\pm$ 0.004       &  0.688 $\pm$ 0.009      &  0.695 $\pm$ 0.004       & 0.759 $\pm$ 0.002  &  0.756 $\pm$ 0.003      & 0.763 $\pm$ 0.001 & 0.745 $\pm$ 0.003   \\ \hline
GAT w/o TopoEdge             & 0.819 $\pm$ 0.010       &   0.840 $\pm$ 0.003       &  0.612 $\pm$ 0.030      &   0.623 $\pm$ 0.035       & 0.742 $\pm$ 0.007  &  0.751 $\pm$ 0.005      & 0.702 $\pm$ 0.018 & 0.716 $\pm$ 0.005     \\
GAT w/ TopoEdge           & 0.861 $\pm$ 0.014       &   0.858 $\pm$ 0.002       &  0.641 $\pm$ 0.021      &  0.646 $\pm$ 0.022        & 0.744 $\pm$ 0.007  &  0.753 $\pm$ 0.005       & 0.758 $\pm$ 0.008 & 0.739 $\pm$ 0.002 \\ \hline
ChebNet w/o TopoEdge         & 0.803 $\pm$ 0.002       &   0.830 $\pm$ 0.002       &  0.530 $\pm$ 0.010      &  0.507 $\pm$ 0.026        & 0.751 $\pm$ 0.009  & 0.771 $\pm$ 0.002        & 0.707 $\pm$ 0.010 &  0.728 $\pm$ 0.006   \\
ChebNet w/ TopoEdge      & 0.865 $\pm$ 0.006       &   0.865 $\pm$ 0.006       &  0.695 $\pm$ 0.002      &   0.693 $\pm$ 0.005       & 0.782 $\pm$ 0.009  &  0.783 $\pm$ 0.001       & 0.758 $\pm$ 0.008  &  0.746 $\pm$ 0.003 \\ \hline
\end{tabular}
\end{table*}

\begin{table*}[t]
\footnotesize 
\vspace{-0.75ex}
\setlength\tabcolsep{5pt}
\caption{Multi-class classification results 
suggest that applying our proposed TopoEdge can achieve steady improvement  
(for both balanced accuracy and Macro-F1)
with various GNN backbones over SOTA baselines.}
\label{result1}
\vskip -2ex
\begin{tabular}{|c|cc|cc|cc|}
\cline{2-7}
\multicolumn{1}{c|}{}  &
  \multicolumn{2}{c|}{Bitcoin-alpha} &
  \multicolumn{2}{c|}{NID} & 
  \multicolumn{2}{c|}{MAG} \\ \cline{2-7}
\multicolumn{1}{c|}{}  & B\_acc & Macro\_F1 & B\_acc & Macro\_F1 & B\_acc & Macro\_F1   \\ \hline
ExtWF         & 0.250 $\pm$ 0.000 & 0.237 $\pm$ 0.000 & 0.238 $\pm$ 0.000  &  0.246 $\pm$ 0.000 & 0.197 $\pm$ 0.000 & 0.175 $\pm$ 0.000 \\ \hline
TER+AER (Geometric)                & 0.348 $\pm$ 0.060 & 0.329 $\pm$ 0.057 & 0.270 $\pm$ 0.071 & 0.167 $\pm$ 0.063 & 0.541 $\pm$ 0.031 & 0.647 $\pm$ 0.050   \\
TER+AER (Possion)           & 0.381 $\pm$ 0.100 & 0.352 $\pm$ 0.100 & 0.316 $\pm$ 0.006 & 0.242 $\pm$ 0.009 & 0.534 $\pm$ 0.026 & 0.627 $\pm$ 0.039   \\ \hline
GCN w/o TopoEdge              &  0.521 $\pm$ 0.009 & 0.559 $\pm$ 0.013        &  0.490 $\pm$ 0.005      &  0.501 $\pm$ 0.008   & 0.570 $\pm$ 0.003 &  0.606 $\pm$ 0.007       \\
GCN w/ TopoEdge          &  0.558 $\pm$ 0.005 & 0.559 $\pm$ 0.005         &  0.516 $\pm$ 0.025      &  0.503 $\pm$ 0.004    & 0.619 $\pm$ 0.009 &  0.623 $\pm$ 0.009       \\ \hline
GAT w/o TopoEdge             &  0.517 $\pm$ 0.012 & 0.558 $\pm$ 0.011         &  0.479 $\pm$ 0.010      &  0.462 $\pm$ 0.014     & 0.601 $\pm$ 0.010 &  0.640 $\pm$ 0.008       \\
GAT w/ TopoEdge          &  0.556 $\pm$ 0.001      &  0.583 $\pm$ 0.001        &  0.528 $\pm$ 0.030      &   0.490 $\pm$ 0.010    & 0.678 $\pm$ 0.007 &  0.652 $\pm$ 0.004      \\ \hline
ChebNet w/o TopoEdge         &  0.514 $\pm$ 0.004      &  0.547 $\pm$ 0.007        &  0.532 $\pm$ 0.019      &   0.510 $\pm$ 0.007      & 0.624 $\pm$ 0.008 &  0.658 $\pm$ 0.004     \\
ChebNet w/ TopoEdge       &  0.542 $\pm$ 0.009      &  0.567 $\pm$ 0.010        &  0.556 $\pm$ 0.014      &   0.534 $\pm$ 0.006      & 0.690 $\pm$ 0.006 &  0.698 $\pm$ 0.001    \\ \hline
\end{tabular}
\end{table*}

\subsection{Experiment Setting}

Our training, validation, and testing sets, were set to 1:2:2 for all datasets. ExtWF~\cite{aggarwal2016edge} and TER+AER~\cite{wang2023efficient} are selected as baselines. ExtWF is a classic heuristic method that performs edge classification based on the dominant labels in similar edge sets, and TER+AER is a newly proposed edge classification method that is based on both high-order proximities in local subgraphs and edge attributes. GCN~\cite{kipf2016semi}, GAT~\cite{velivckovic2017graph}, and ChebNet~\cite{tang2019chebnet} are selected as the backbones. We follow the setting in \cite{huang2021sdgnn}, where we first use these methods to get node representations. For edge $e_{ij}$, we concatenate these two learned end node representations $z_i$ and $z_j$ to compose an edge representation $z_{ij}$. After that, we train an MLP classifier on the training set and use it to predict the edge label in the testing set. As previous empirical studies do, balanced accuracy and Macro-F1 are selected as the evaluation metrics, which gives insight into the model's performance over different classes. The temperature hyperparameter \( t \) is finely tuned within the range \((0, 5]\). The parameter \( K \) is adjusted from \(0\) up to the size of the training data, while both \( \theta \) and \( h \) are tuned within the interval \((0,1)\).

\vspace{-1ex}
\subsection{Classification Performance Evaluation}

The classification results are shown in Table~\ref{fig-binary} and Table~\ref{result1}. Overall the proposed TopoEdge depicts significant improvement against backbones. Also, applying our strategy on backbones that are not explicitly designed for learning edge representation can achieve a similar level of performance with the most advanced baseline (TER+AER) in binary classification tasks and significantly beat it in multi-class classification. This suggests the effectiveness of our proposed strategies rather than empathizing edge with high distribution variance can mitigate the topological imbalance issues. The balanced accuracy and Macro-F1 score in binary classification is higher than that in multi-class classification, which is because the edges are influenced by more classes during message passing and the aggregated message is of relatively low quality. This again shows the need to address topological imbalance in edge classification. ExtWF performs poorly across datasets because it is based on the dominant edge class in similar edge sets, edges will always tend to be predicted as the majority of edges in a highly imbalanced dataset. TER+AER can depict comparable performance in binary classification as it involves generating edge embedding based on higher-order proximity in local subgraphs, which shows that leveraging local structural patterns can indeed boost performance. However, as this method does not consider the influence of quantity and topological imbalance, hence when uncertainty in the local subgraph arises by switching to multi-class datasets, its performance degrades a lot. This also suggests that compared with focusing on local proximity, emphasizing the variance will help increase the edge classification performance more on the imbalanced dataset.

\subsection{Ablation Study}\label{sec-ablation}

\begin{figure}[t]

     \centering
     \begin{subfigure}[b]{0.21\textwidth}
         \centering
         \includegraphics[width=\textwidth]{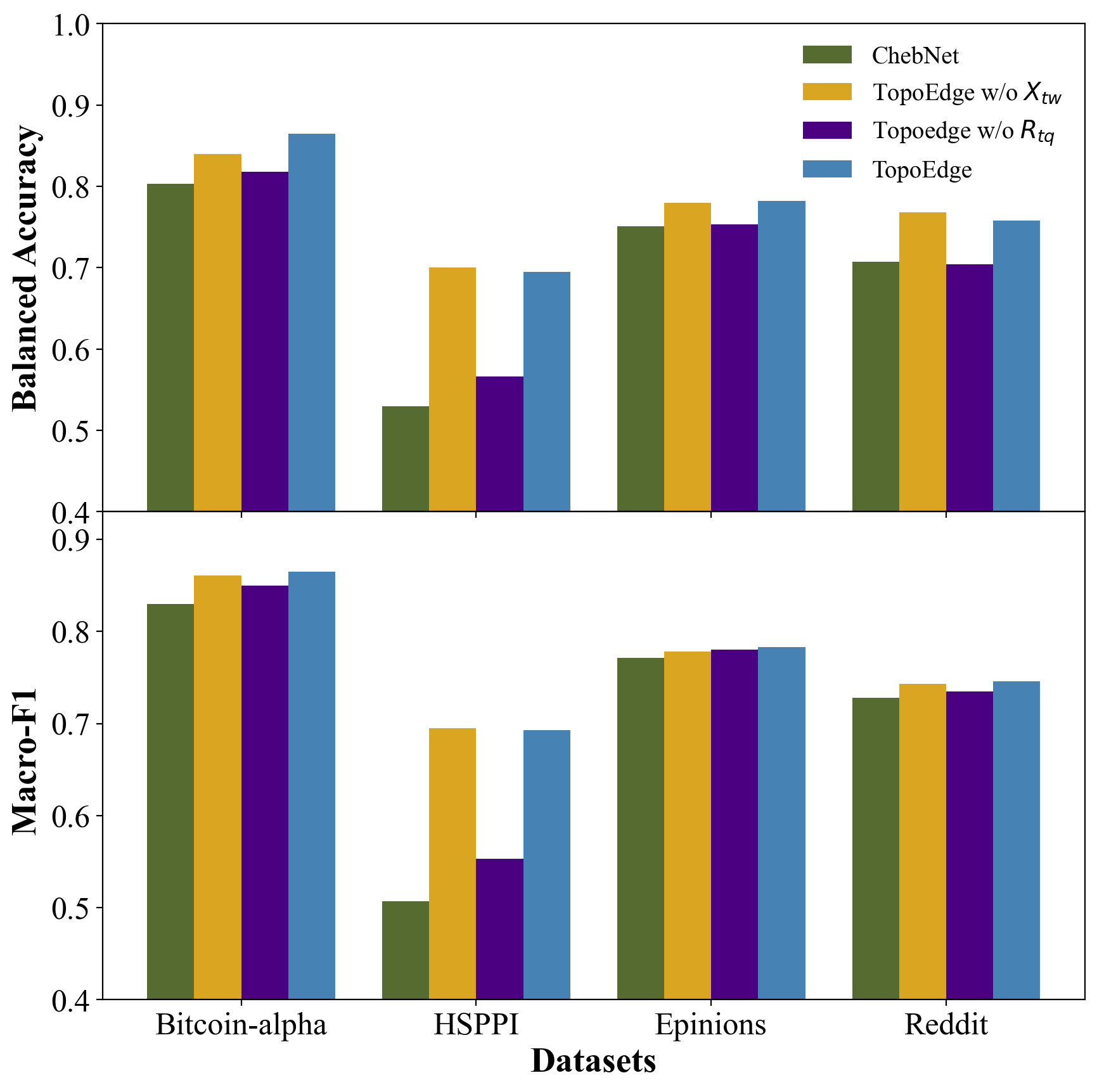}
         \caption{}
         \label{fig-abl1}
     \end{subfigure}
     \hspace{5ex}
     \begin{subfigure}[b]{0.21\textwidth}
         \centering
         \includegraphics[width=\textwidth]{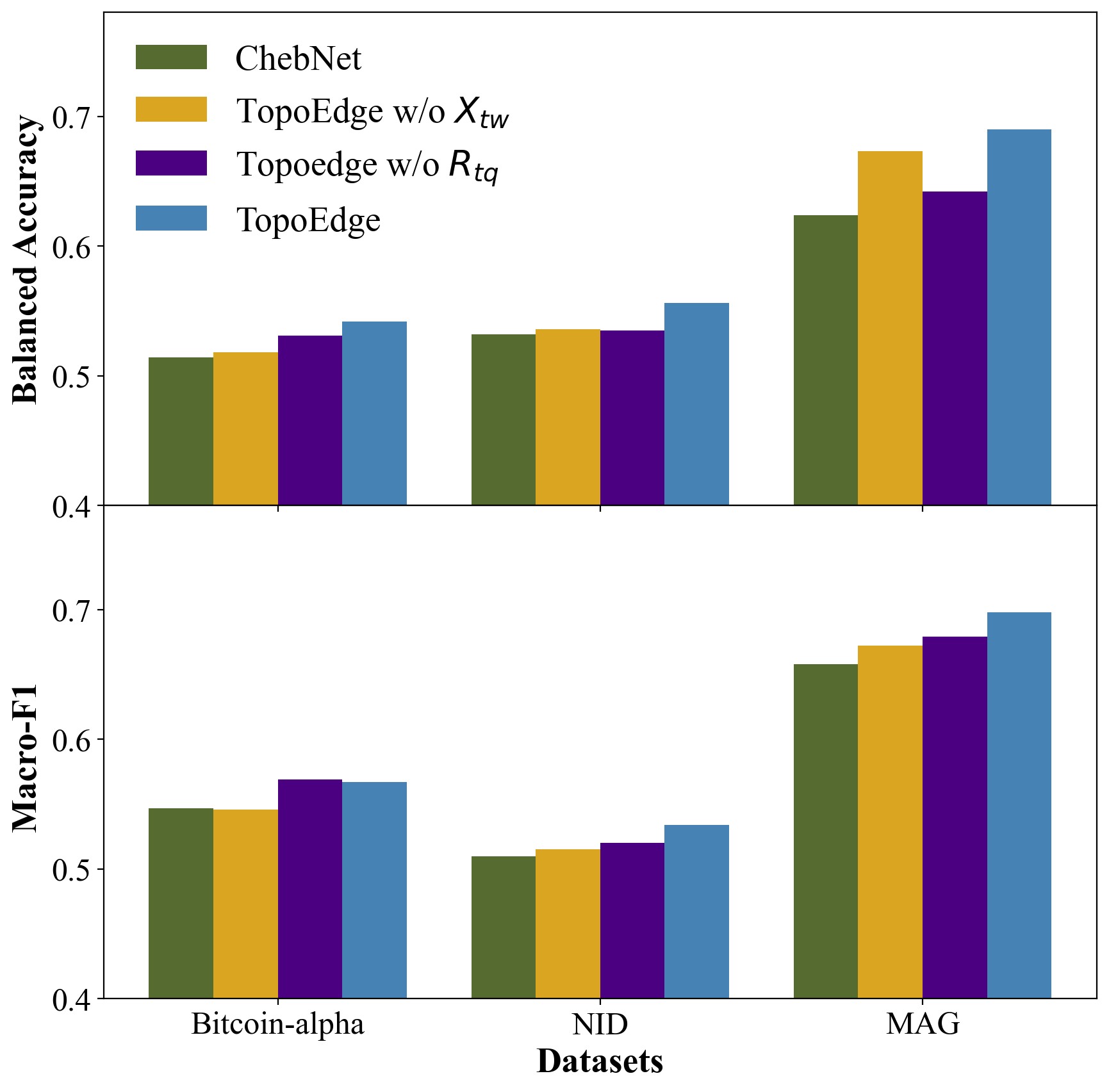}
         \caption{}
         \label{fig-abl2}
     \end{subfigure}
     \vskip -1.5ex
    \caption{Ablation studies for edge classification tasks, with \textbf{(a)} showing the performance on binary classification tasks, and \textbf{(b)} showing the results for multi-class classifications. Overall, TopoEdge depicts the best performance, and topological reweight and TE wedge-based mixup can independently improve the performance.}
     \label{fig-ablation}
     \vskip -3.5ex
\end{figure}

In this section, we detail the ablation studies conducted to evaluate the impact of different components of the TopoEdge framework on edge classification tasks. The results, as illustrated in Figure~\ref{fig-ablation}, employing the full TopoEdge framework significantly enhances classification performance compared to using these components separately or the baseline models alone in both tasks. More specifically, the integration of both components in TopoEdge yields the most substantial improvements, demonstrating a synergistic effect that optimizes classification outcomes. Also, both Topological Reweight and TE Wedge-Based Mixup independently improve performance over baseline models, confirming their efficacy. These findings underscore the effectiveness of each component and highlight the benefits of their integration for imbalanced edge classification.

\subsection{Influence of Labeled Data Ratio}\label{sec-low-label}

\begin{figure}[t]
     \centering
     \begin{subfigure}[b]{0.23\textwidth}
         \centering
         \includegraphics[width=\textwidth]{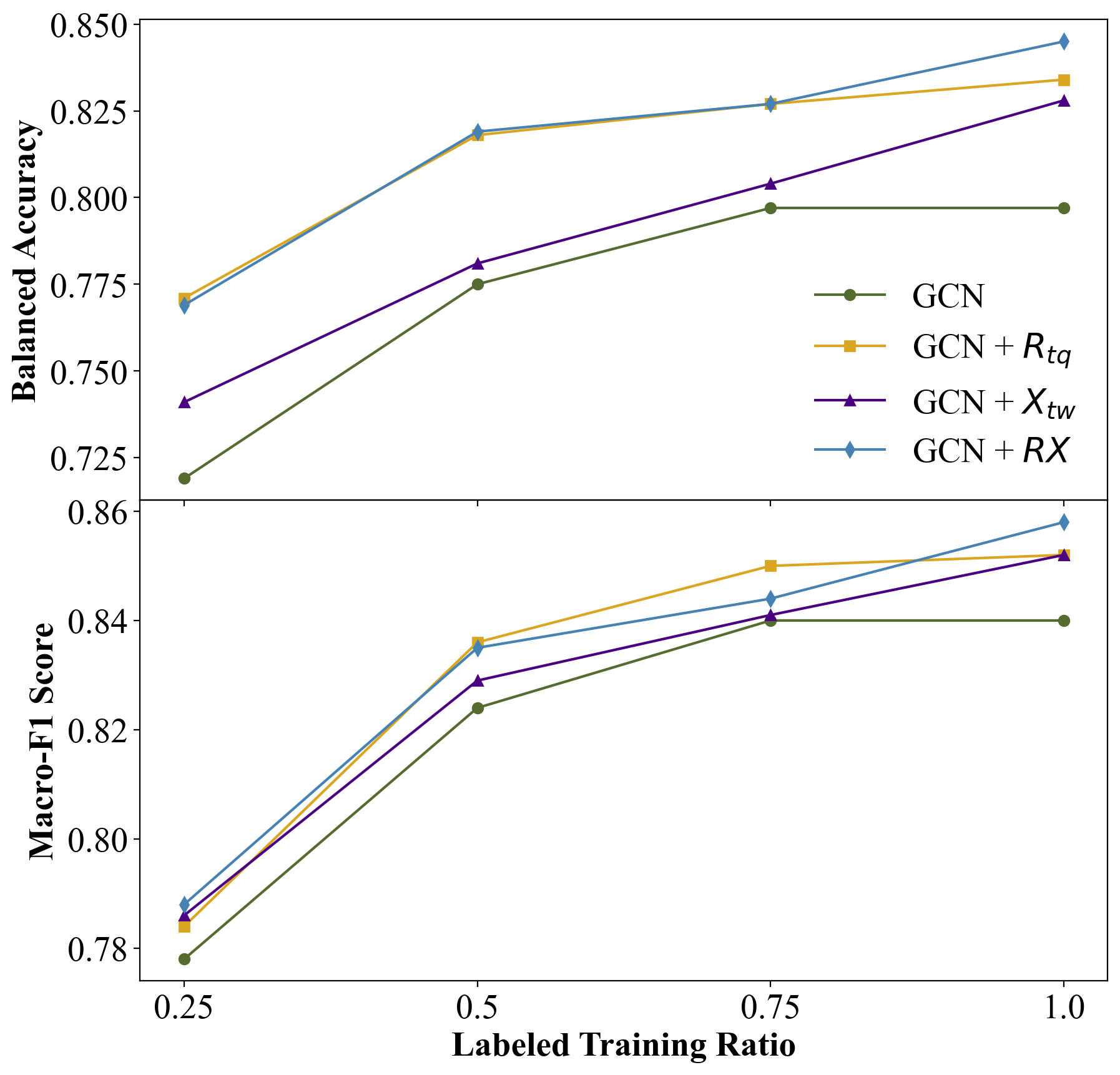}
         \caption{Bitcoin-alpha (multi-class)}
         \label{fig-label-ba}
     \end{subfigure}
     \begin{subfigure}[b]{0.23\textwidth}
         \centering
         \includegraphics[width=\textwidth]{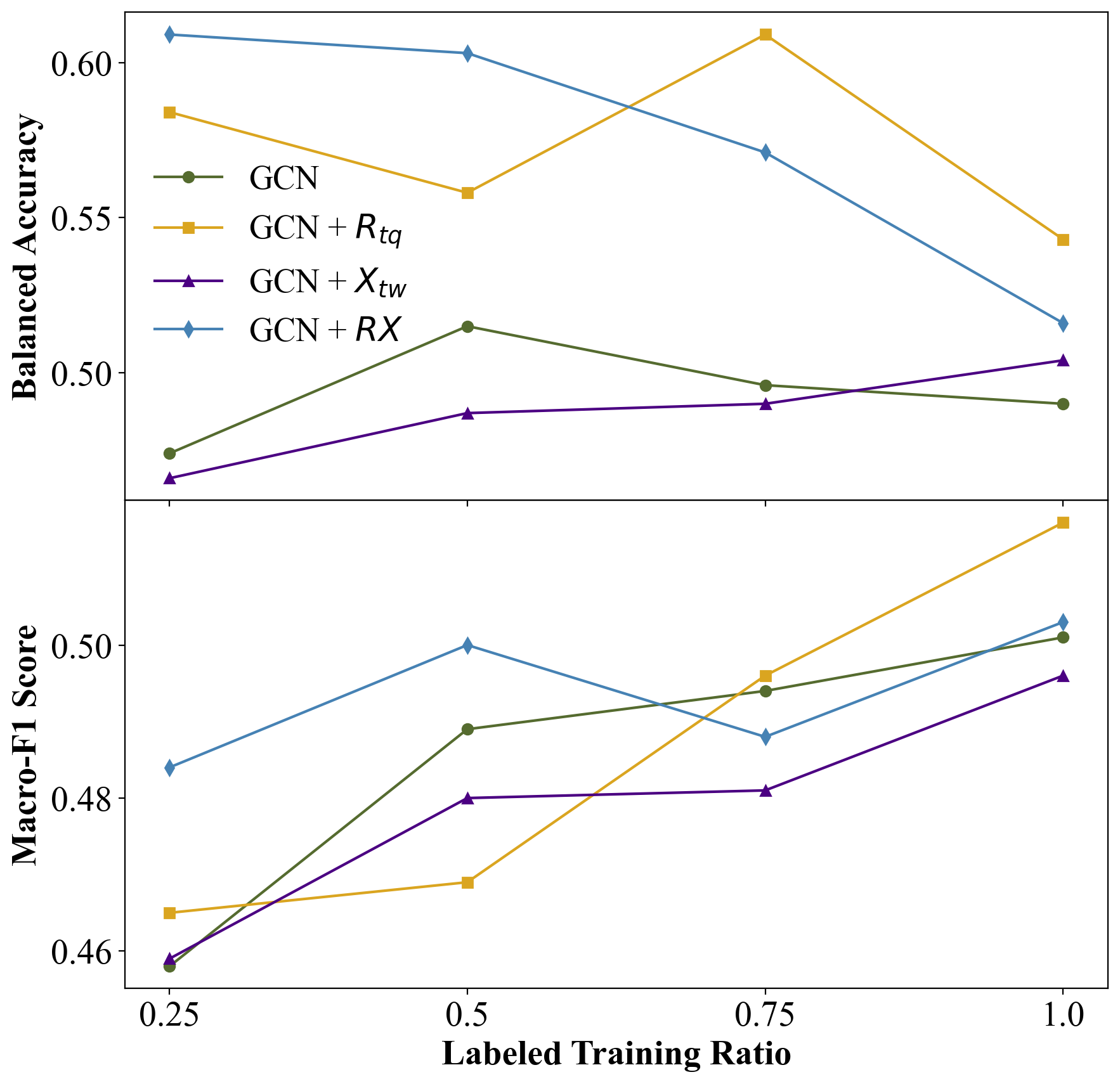}
         \caption{NID}
         \label{fig-label-int}
     \end{subfigure}
     \vskip -1.5ex
     \caption{The classification performance of \textbf{(a)} Bitcoin-alpha (multi-class) and \textbf{(b)} NID under various labeled training ratios (e.g., 0.5 means only leveraging half of the training dataset)}
     \label{fig-label}
     \vskip -1ex
\end{figure}

Generally speaking, the labeling ratios in graph datasets are low, so in this section, we give a further analysis of the classification performance of our proposed method under various labeled training ratios. As shown in Figure~\ref{fig-label}, our proposed strategies can steadily beat the baseline even when low-labeled training data is available, and TopoEdge (+ $RX$ ) depicts the best performance under an extremely low labeled ratio. This indicates that capturing local distribution variance works even when scare data is available. as scare data may still depict large variance and we can take advantage of that to get better edge representation.

\vspace{-1.5ex}
\section{Related Work}\label{sec-relatedwork}
In this section, we present a summary of the related work in edge classification and imbalanced graph machine learning, which are the two closest directions to the presented work. 

\vspace{1ex}

\textbf{Edge Classification: }
The problem of edge classification in networks was formulated by Aggarwal et al.~\cite{aggarwal2016edge} in 2016, where the authors address the edge classification tasks by proposing a structural similarity model using the weighted Jaccard coefficient as the underlying structural proximity metric. In recent years, most work in the field of edge classification has focused on generating embeddings for edges. One direction uses shallow embedding techniques. For example, AttrE2vec~\cite{bielak2022attre2vec} aggregates features through random walk, Edge2vec~\cite{wang2020edge2vec} uses deep auto-encoders, and TER+AER~\cite{wang2023efficient} generate edge embeddings by encoding local topology structure information while augmenting edge attributes through a feature aggregation scheme. Another direction lies in GNNs, where the most commonly seen approach is to use the end nodes' embeddings and form edge embeddings via operations such as averaging, Hadamard product, concatenation, or deep neural network~\cite{kipf2016semi,velivckovic2017graph, tang2019chebnet}. Converting the original graphs into line graphs and then using GNNs to generate edge representation has also been studied~\cite{wang2020generic}. What's more, even though methods like EGNN~\cite{kim2019edge} and EGAT~\cite{gong2019exploiting} are used for node classification, they can generate embeddings for edges directly, without first converting edges into nodes. However,  none of these above methods focused on mitigating the topological imbalance in edge classification.

\vspace{1ex}

\textbf{Imbalanced Graph Machine Learning: }
Imbalanced graph machine learning is an emerging research area addressing imbalance issues in graph data. In recent years, several methods have been proposed~\cite{chen2021topology, liu2021pick, park2021graphens, zhao2021graphsmote, liu2023topological, wang2022imbalanced, sun2022position} in this field. One of the early works GraphSMOTE ~\cite{zhao2021graphsmote} adopts SMOTE~\cite{chawla2002smote} oversampling in the node embedding space to synthesize minority nodes and complements the topology with a learnable edge predictor. In the field of topology imbalance, ReNode~\cite{chen2021topology} first addresses the topology imbalance issue, i.e., the unequal structure role of labeled nodes in the topology, by down-weighting close-to-boundary labeled nodes. PASTEL~\cite{sun2022position} addresses the under-reaching and over-squashing issue in node classification through structural learning. TOBA ~\cite{liu2023topological} approaches the source of the class-imbalance bias from a topology-centric perspective. However, none of the existing research focused on an edge-centric perspective. To the best of our knowledge, there are currently no extensive studies related to topology imbalance issues in edge classification. In this paper, we approach the topology imbalance issue in edge classification using a reweighting and resampling method, and consequently future boost the current benchmark performance on edge classification.

\section{Conclusion}\label{sec-conclusion}

In this paper, we addressed the previously overlooked issue of topological imbalance in edge classification within graph machine learning (GML). We found that this imbalance significantly hinders model performance across various graphs. To measure these imbalances, we introduced a new metric, Topological Entropy (TE), which quantifies class distribution variance in local subgraphs. Building on this, we developed a topological reweighting approach to adjust the training weights of labeled edges based on their topological properties. Additionally, we introduced a topological wedge-based mixup strategy that generates synthetic training edges through interpolation among high-TE edges. These strategies are integrated into our novel TopoEdge approach for edge classification. Our comprehensive empirical analyses confirm the effectiveness of TopoEdge in various classification tasks. Acknowledging the importance of topological imbalance, we anticipate further dedicated research in this direction. Given the widespread nature of topological imbalances, exploring resolutions across more applications is a promising direction for future research.

\balance 
\bibliographystyle{ACM-Reference-Format}
\bibliography{reference}
\balance 

\end{document}